\documentclass[10pt,journal,compsoc]{IEEEtran}
\ifCLASSOPTIONcompsoc
  \usepackage[nocompress]{cite}
\else
  \usepackage{cite}
\fi
\ifCLASSINFOpdf
  \usepackage[pdftex]{graphicx}
  \graphicspath{{../pdf/}{../jpeg/}}
  \DeclareGraphicsExtensions{.pdf,.jpeg,.png}
\else
  \usepackage[dvips]{graphicx}
  \graphicspath{{../eps/}}
  \DeclareGraphicsExtensions{.eps}
\fi

\usepackage{amsmath}
\ifCLASSOPTIONcompsoc
 \usepackage[caption=false,font=footnotesize,labelfont=sf,textfont=sf]{subfig}
\else
 \usepackage[caption=false,font=footnotesize]{subfig}
\fi
\usepackage{amssymb}
\usepackage{pifont}
\newcommand{\cmark}{\ding{51}}
\newcommand{\xmark}{\ding{55}}
\usepackage{bm}

\usepackage{ragged2e} %zj
\usepackage{url} %zj
\usepackage[breaklinks,colorlinks]{hyperref}
\hypersetup{colorlinks=true}

\usepackage{tikz}
\usetikzlibrary{
    arrows.meta,
    shapes.geometric,
    backgrounds,
    fit,
    calc,
    positioning
}
\tikzset{
    section/.style={
        shape=rectangle, rounded corners, fill=#1!30,
        minimum width=4cm,
        minimum height=0.8cm,
        text width=4.8cm,
        text centered
    },
    every label/.append style={font=\scriptsize},
    box/.style={
        shape=rectangle, rounded corners, draw=gray, dashed, thick, inner sep=3pt
    },
    process/.style={
        shape=rectangle, rounded corners,
        draw,
        minimum width=2cm,
        minimum height=0.5cm,
        text width=2.2cm,
        text centered
    },
    line/.style={ultra thick, gray},
    flow line/.style={-latex}
}
\usepackage{subfiles}

\begin{document}
%
% paper title
% Titles are generally capitalized except for words such as a, an, and, as,
% at, but, by, for, in, nor, of, on, or, the, to and up, which are usually
% not capitalized unless they are the first or last word of the title.
% Linebreaks \\ can be used within to get better formatting as desired.
% Do not put math or special symbols in the title.
\title{Event-based Simultaneous Localization and Mapping: A Comprehensive Survey}
%
%
% author names and IEEE memberships
% note positions of commas and nonbreaking spaces ( ~ ) LaTeX will not break
% a structure at a ~ so this keeps an author's name from being broken across
% two lines.
% use \thanks{} to gain access to the first footnote area
% a separate \thanks must be used for each paragraph as LaTeX2e's \thanks
% was not built to handle multiple paragraphs
%
%
%\IEEEcompsocitemizethanks is a special \thanks that produces the bulleted
% lists the Computer Society journals use for "first footnote" author
% affiliations. Use \IEEEcompsocthanksitem which works much like \item
% for each affiliation group. When not in compsoc mode,
% \IEEEcompsocitemizethanks becomes like \thanks and
% \IEEEcompsocthanksitem becomes a line break with idention. This
% facilitates dual compilation, although admittedly the differences in the
% desired content of \author between the different types of papers makes a
% one-size-fits-all approach a daunting prospect. For instance, compsoc 
% journal papers have the author affiliations above the "Manuscript
% received ..."  text while in non-compsoc journals this is reversed. Sigh.

\author{Kunping~Huang,
        Sen Zhang, %~\IEEEmembership{,}
        Jing Zhang,~\IEEEmembership{Senior Member,~IEEE},
        and Dacheng Tao,~\IEEEmembership{Fellow,~IEEE}
        % <-this % stops a space
\IEEEcompsocitemizethanks{\IEEEcompsocthanksitem The authors are with the School of Computer Science, Faculty of
Engineering, University of Sydney, Darlington, NSW 2008\protect\\
% note need leading \protect in front of \\ to get a newline within \thanks as
% \\ is fragile and will error, could use \hfil\break instead.
E-mail: \{khua7609, szha2609\}@uni.sydney.edu.au; \{jing.zhang1, dacheng.tao\}@sydney.edu.au}% <-this % stops an unwanted space
%\thanks{Manuscript received April 19, 2005; revised August 26, 2015.}
}

% note the % following the last \IEEEmembership and also \thanks - 
% these prevent an unwanted space from occurring between the last author name
% and the end of the author line. i.e., if you had this:
% 
% \author{....lastname \thanks{...} \thanks{...} }
%                     ^------------^------------^----Do not want these spaces!
%
% a space would be appended to the last name and could cause every name on that
% line to be shifted left slightly. This is one of those "LaTeX things". For
% instance, "\textbf{A} \textbf{B}" will typeset as "A B" not "AB". To get
% "AB" then you have to do: "\textbf{A}\textbf{B}"
% \thanks is no different in this regard, so shield the last } of each \thanks
% that ends a line with a % and do not let a space in before the next \thanks.
% Spaces after \IEEEmembership other than the last one are OK (and needed) as
% you are supposed to have spaces between the names. For what it is worth,
% this is a minor point as most people would not even notice if the said evil
% space somehow managed to creep in.

% The paper headers
\markboth{Journal of \LaTeX\ Class Files,~Vol.~14, No.~8, August~2015}%
{Huang \MakeLowercase{\textit{et al.}}: Event-based Simultaneous Localization and Mapping: A Comprehensive Survey}
% The only time the second header will appear is for the odd numbered pages
% after the title page when using the twoside option.
% 
% *** Note that you probably will NOT want to include the author's ***
% *** name in the headers of peer review papers.                   ***
% You can use \ifCLASSOPTIONpeerreview for conditional compilation here if
% you desire.

% The publisher's ID mark at the bottom of the page is less important with
% Computer Society journal papers as those publications place the marks
% outside of the main text columns and, therefore, unlike regular IEEE
% journals, the available text space is not reduced by their presence.
% If you want to put a publisher's ID mark on the page you can do it like
% this:
%\IEEEpubid{0000--0000/00\$00.00~\copyright~2015 IEEE}
% or like this to get the Computer Society new two part style.
%\IEEEpubid{\makebox[\columnwidth]{\hfill 0000--0000/00/\$00.00~\copyright~2015 IEEE}%
%\hspace{\columnsep}\makebox[\columnwidth]{Published by the IEEE Computer Society\hfill}}
% Remember, if you use this you must call \IEEEpubidadjcol in the second
% column for its text to clear the IEEEpubid mark (Computer Society jorunal
% papers don't need this extra clearance.)

% use for special paper notices
%\IEEEspecialpapernotice{(Invited Paper)}

% for Computer Society papers, we must declare the abstract and index terms
% PRIOR to the title within the \IEEEtitleabstractindextext IEEEtran
% command as these need to go into the title area created by \maketitle.
% As a general rule, do not put math, special symbols or citations
% in the abstract or keywords.
\IEEEtitleabstractindextext{%
\begin{abstract}
\justifying
In recent decades, visual simultaneous localization and mapping (vSLAM) has gained significant interest in both academia and industry. It estimates camera motion and reconstructs the environment concurrently using visual sensors on a moving robot. However, conventional cameras are limited by hardware, including motion blur and low dynamic range, which can negatively impact performance in challenging scenarios like high-speed motion and high dynamic range illumination. Recent studies have demonstrated that event cameras, a new type of bio-inspired visual sensor, offer advantages such as high temporal resolution, dynamic range, low power consumption, and low latency. This paper presents a timely and comprehensive review of event-based vSLAM algorithms that exploit the benefits of asynchronous and irregular event streams for localization and mapping tasks. The review covers the working principle of event cameras and various event representations for preprocessing event data. It also categorizes event-based vSLAM methods into four main categories: feature-based, direct, motion-compensation, and deep learning methods, with detailed discussions and practical guidance for each approach. Furthermore, the paper evaluates the state-of-the-art methods on various benchmarks, highlighting current challenges and future opportunities in this emerging research area. A public repository will be maintained to keep track of the rapid developments in this field at {\url{https://github.com/kun150kun/ESLAM-survey}}.
\end{abstract}

% Note that keywords are not normally used for peerreview papers.
\begin{IEEEkeywords}
Event camera, visual SLAM, visual odometry, 3D reconstruction, Deep learning
\end{IEEEkeywords}}

% make the title area
\maketitle

% To allow for easy dual compilation without having to reenter the
% abstract/keywords data, the \IEEEtitleabstractindextext text will
% not be used in maketitle, but will appear (i.e., to be "transported")
% here as \IEEEdisplaynontitleabstractindextext when the compsoc 
% or transmag modes are not selected <OR> if conference mode is selected 
% - because all conference papers position the abstract like regular
% papers do.
\IEEEdisplaynontitleabstractindextext
% \IEEEdisplaynontitleabstractindextext has no effect when using
% compsoc or transmag under a non-conference mode.

% For peer review papers, you can put extra information on the cover
% page as needed:
% \ifCLASSOPTIONpeerreview
% \begin{center} \bfseries EDICS Category: 3-BBND \end{center}
% \fi
%
% For peerreview papers, this IEEEtran command inserts a page break and
% creates the second title. It will be ignored for other modes.
\IEEEpeerreviewmaketitle

\IEEEraisesectionheading{\section{Introduction}\label{sec:introduction}}
% Computer Society journal (but not conference!) papers do something unusual
% with the very first section heading (almost always called "Introduction").
% They place it ABOVE the main text! IEEEtran.cls does not automatically do
% this for you, but you can achieve this effect with the provided
% \IEEEraisesectionheading{} command. Note the need to keep any \label that
% is to refer to the section immediately after \section in the above as
% \IEEEraisesectionheading puts \section within a raised box.

% The very first letter is a 2 line initial drop letter followed
% by the rest of the first word in caps (small caps for compsoc).
% 
% form to use if the first word consists of a single letter:
% \IEEEPARstart{A}{demo} file is ....
% 
% form to use if you need the single drop letter followed by
% normal text (unknown if ever used by the IEEE):
% \IEEEPARstart{A}{}demo file is ....
% 
% Some journals put the first two words in caps:
% \IEEEPARstart{T}{his demo} file is ....
% 
% Here we have the typical use of a "T" for an initial drop letter
% and "HIS" in caps to complete the first word.
\IEEEPARstart{V}{isual} simultaneous localization and mapping (vSLAM), which concurrently estimates robot localization and reconstructs a 3D map of the surrounding environment using onboard visual sensors, has become an essential component for various applications~\cite{7747236,robotics11010024,zhang2020empowering}, including autonomous driving, robot navigation, and augmented reality. For instance, to navigate to designated positions and accomplish various tasks, robots require information on the environment and their internal states, which can be provided by vSLAM algorithms. Despite the benefits of vSLAM, most algorithms rely on conventional frame-based cameras that capture full-brightness intensity images within a short exposure time at a fixed frequency. However, frame-based cameras suffer from several drawbacks. First, they acquire intensity images within an exposure time window, which can result in blurred images due to a sharp brightness change when either the camera or the objects in the scene are moving rapidly. Moreover, a long exposure time window in a night scene to capture more light will also lead to blurry images caused by the moving camera or objects as shown in \cite{7169560, 9578515, Liu2021MBAVOMB}. Consequently, current vSLAM algorithms tend to lose camera tracking when they receive these low-quality blurry images. Second, frame-based cameras cannot capture accurate and informative intensity information under extreme brightness or darkness in the environment due to their low dynamic range, leading to vSLAM system failures in illumination-challenging environments. Third, the frame-based camera obtains static appearance-based features at a fixed rate, which requires an additional frame interpolation step to track temporal dynamics \cite{10089678} for moving objects.

\begin{figure}
    \centering
    \subfloat[Raw event data in the spatio-temporal space]{
        \includegraphics[width=0.23\textwidth]{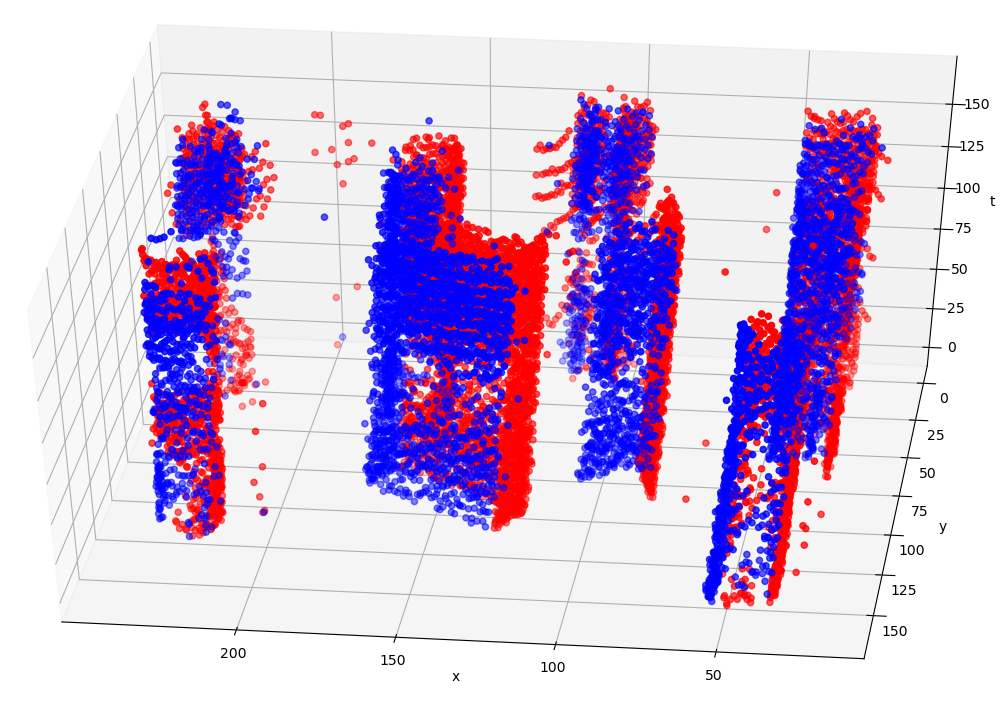}
        \label{subfig:raw_event}
    }
    \hfil
    \subfloat[Event data on the image plane]{
        \includegraphics[width=0.2\textwidth]{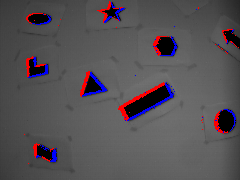}
        \label{subfig:event_image}
    }
    \hfil
    \subfloat[Event-based vSLAM system \cite{9879881}]{
        \includegraphics[width=0.45\textwidth]{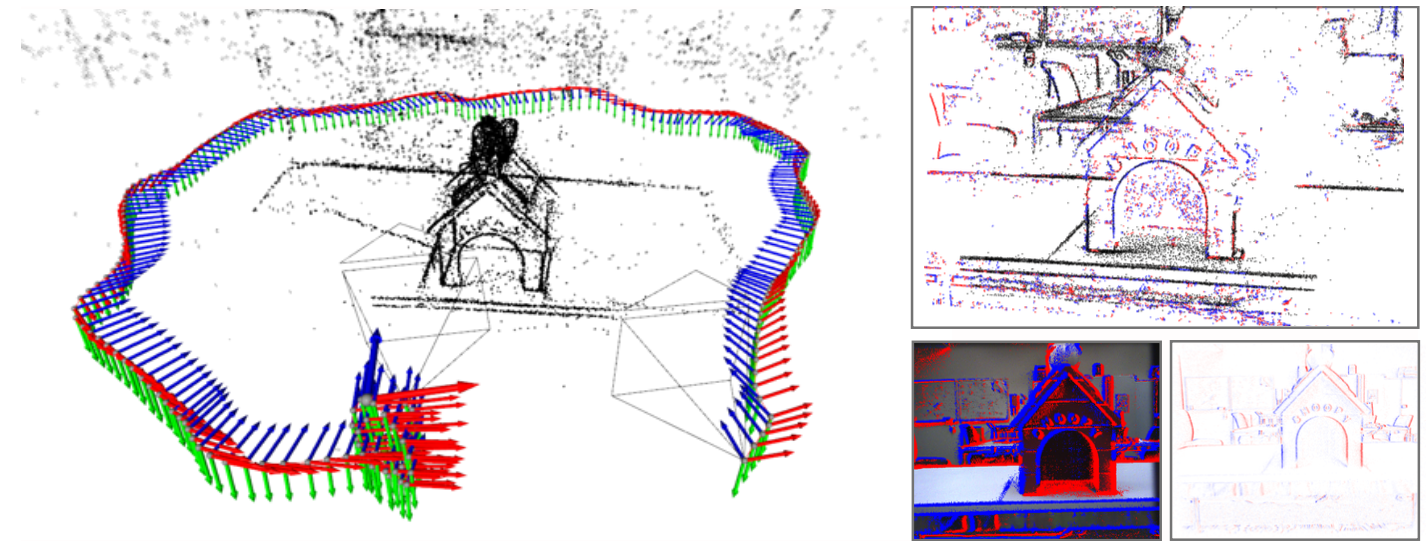}
        \label{subfig:event_slam}
    }
    \caption{Figure (a) and (b) are raw event data in the spatio-temporal space and projected on the image plane, respectively with pseduo-colored red-blue points, according to event polarity. Figure (c), adapted from \cite{9879881}, shows the camera trajectory and 3D depth map.}
    \label{fig:event}
\end{figure}

In recent years, \textit{event camera}, a bio-inspired sensor, has attracted extensive research interest for its potential to overcome the intrinsic problems of frame-based cameras for vSLAM tasks in challenging environments characterized by high-speed motion and high dynamic range (HDR) lighting conditions. Unlike conventional frame-based cameras, the event camera operates in a fundamentally different way. It records brightness change (called \textit{event}) at each independent pixel and outputs event data asynchronously (Fig.~\ref{subfig:raw_event}). Compared to frame-based cameras, event cameras provide four favorable advantages \cite{Gallego2019EventBasedVA}: 1) event cameras have a high temporal resolution, enabling them to monitor intensity changes and output events in microseconds,  facilitating their adaptation to high-speed motion without the issue of motion blur and tracking temporal dynamics cues \cite{10089678} for the moving object contours and textures; 2) since each pixel operates independently and event data is transmitted immediately upon the occurrence of a sufficient change, event cameras have very low latency, which allows vSLAM algorithms to process event data in continuous time and track robot states continuously to estimate smooth camera motion; 3) event cameras consume less power than most frame-based cameras \cite{Gallego2019EventBasedVA, 8100264, 9811805, 6906882} due to their low-redundancy data transmission and processing, making them suitable for resource-limited systems; and 4) the logarithmic scale of brightness change captured by event cameras allows for a remarkably higher dynamic range ($>120$ dB) compared to conventional frame-based cameras ($<60$ dB), enabling them to be applicable from day to night.

\begin{figure*}
    \centering
    \resizebox{1.7\columnwidth}{!}{
    \begin{tikzpicture}[
        node distance=1cm
    ]
        \node[anchor=north, font=\bfseries] (pre) {Preliminary};
        \node[section=yellow, below of=pre, label=Sec. \ref{subsec:work_principle}] (wp) {Working Principle};
        \node[section=yellow, below of=wp, yshift=-0.15cm, label=Sec. \ref{subsec:event_representation}] (er) {Event Representation};
        \node[section=yellow, below of=er, yshift=-0.15cm, label=Sec. \ref{subsec:pipeline}] (pipeline) {Pipeline of vSLAM};
        \begin{scope}[on background layer]
            \node [box, fit=(pre)(wp)(er)(pipeline)] (Prelim) {};
        \end{scope}

        \node[below of=pipeline, yshift=-0.15cm, font=\bfseries] (vo) {Event-based SLAM};
        \node[section=red, below of=vo, yshift=-0.1cm, label={[name=l1]Sec. \ref{sec:feature}}] (feature) {Feature-based Methods};
        \node[section=red, below of=feature, yshift=-0.2cm, label=Sec. \ref{sec:direct}] (direct) {Direct Methods};
        \node[section=red, below of=direct, yshift=-0.2cm, label=Sec. \ref{sec:motion}] (motion) {Motion-compensation Methods};
        \node[section=red, below of=motion, yshift=-0.2cm, label=Sec. \ref{sec:dl}] (dl) {Deep Learning Methods};
        \begin{scope}[on background layer]
            \node[box, fit=(feature)(direct)(motion)(dl)(l1)(vo)] (main) {};
        \end{scope}
        \draw [line, ->] (Prelim) -- (main);

        \node[section=green, right of=er, xshift=5.0cm, label={[name=l2]Sec. \ref{subsec:feature_extraction}}] (f-e) {Feature Extraction};
        \node[section=green, below of=f-e, yshift=-0.2cm, label=Sec. \ref{subsec:feature_tracking}] (f-t) {Feature Tracking};
        \node[section=green, below of=f-t, yshift=-0.2cm, label=Sec. \ref{subsec:feature_tam}] (f-tam) {Camera Tracking and Mapping};
        \begin{scope}[on background layer]
            \node [box, fit=(f-e)(f-t)(f-tam)(l2)] (feature-sub) {};
        \end{scope}
        \draw [line, ->] (feature) -| ($(feature.east)!.5!(feature-sub.west)$) |- (feature-sub);

        \node[section=green, left of=er, xshift=-5.0cm, label={[name=l3]Sec. \ref{subsec:event_image_align}}] (d-f) {Event-Image Alignment};
        \node[section=green, below of=d-f, yshift=-0.2cm, label=Sec. \ref{subsec:event_event_align}] (d-o) {Event Representation-based Alignment};
        \begin{scope}[on background layer]
            \node [box, fit=(d-f)(d-o)(l3)] (direct-sub) {};
        \end{scope}
        \draw [line, ->] (direct) -| ($(direct.west)!.5!(direct-sub.east)$) |- (direct-sub);

        \node[section=green, left of=feature, xshift=-5.0cm, yshift=0.0cm, label={[name=l4]Sec. \ref{subsec:cmax}}] (m-cm) {Contrast Maximization};
        \node[section=green, below of=m-cm, yshift=-0.2cm, label=Sec. \ref{subsec:dmin}] (m-dm) {Dispersion Minimization};
        \node[section=green, below of=m-dm, yshift=-0.2cm, label=Sec. \ref{subsec:prob_model}] (m-pm) {Probabilistic Model};
        \node[section=green, below of=m-pm, yshift=-0.2cm, label=Sec. \ref{subsec:prob_model}] (m-ec) {Event Collapse};
        \begin{scope}[on background layer]
            \node [box, fit=(m-cm)(m-dm)(m-pm)(m-ec)(l4)] (motion-sub) {};
        \end{scope}
        \draw [line, ->] (motion) -| ($(motion.east)!.5!(motion-sub.west)$) |- (motion-sub);

        \node[section=green, right of=motion, xshift=5.0cm, label={[name=l5]Sec. \ref{subsec:dl_unsupervised}}] (dl-e) {Self-supervised Learning};
        \node[section=green, below of=dl-e, yshift=-0.2cm, label=Sec. \ref{subsec:dl_supervised}] (dl-d) {Supervised Learning};
        \begin{scope}[on background layer]
            \node [box, fit=(dl-e)(dl-d)(l5)] (dl-sub) {};
        \end{scope}
        \draw [line, ->] (dl) -| ($(dl.west)!.5!(dl-sub.east)$) |- (dl-sub);

        \node[section=blue, below of=dl, yshift=-1.1cm, xshift=-3.5cm, label={[name=l7]Sec. \ref{sec:perf}}] (perf) {Performance Comparison};
        \node[section=blue, below of=dl, yshift=-1.1cm, xshift=3.5cm, label={[name=l8]Sec. \ref{sec:ch_future}}, text width = 5.5cm] (ch_future) {Challenges and Future Directions};
        \coordinate (main-south) at ($(main.south)+(0,-0.4)$);
        \draw [line, ->] (main) -- (main-south) -| (l7);
        \draw [line, ->] (main) -- (main-south) -| (l8);
    \end{tikzpicture}
    }
    \caption{The structure of the event-based vSLAM algorithms and the taxonomy of the existing works.}
    \label{fig:structure}
\end{figure*}
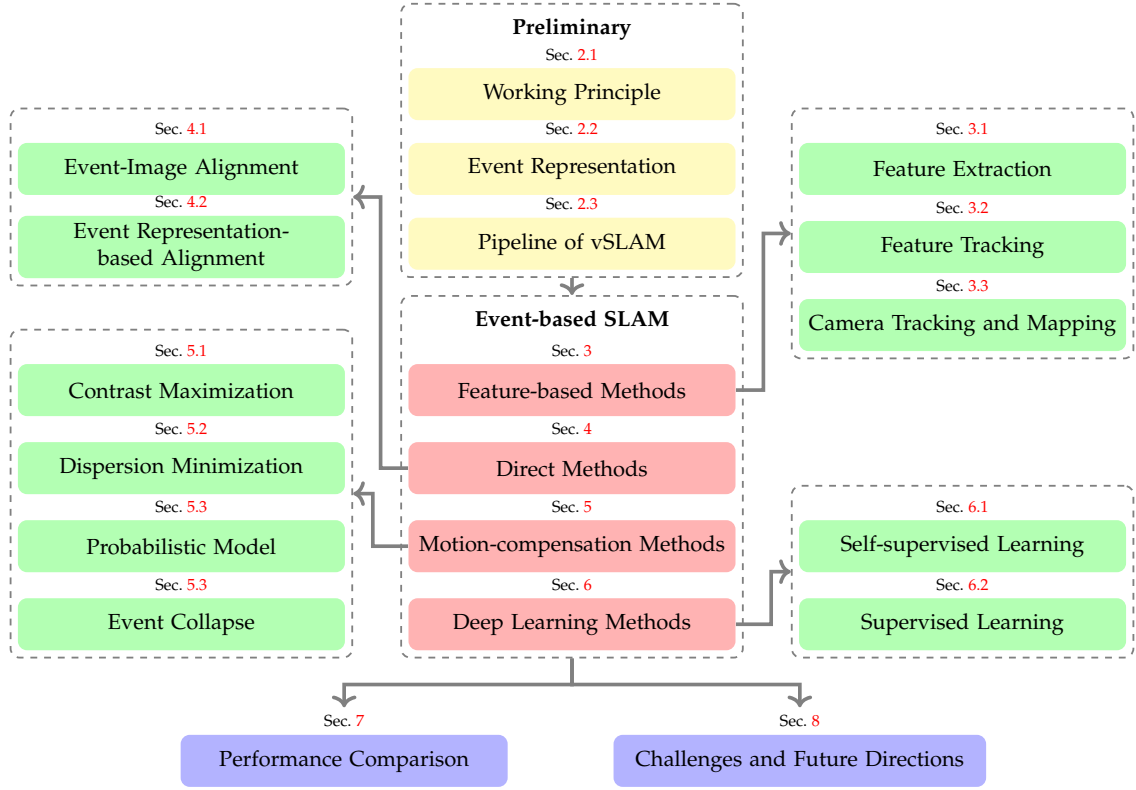

Despite their promising properties in challenging environments, event cameras cannot be directly applied to the existing frame-based vSLAM algorithms that rely on processing 2D dense brightness intensity images \cite{6906584,7946260,7898369}. This is because the event data, which represents the brightness change at each pixel, is time-continuous, sparse, asynchronous, and irregular. Moreover, conventional frame-based vSLAM algorithms utilize intensity information for data association and construct multi-view geometric relationships to jointly estimate camera poses and environmental structures. However, events only convey limited information and contain inherent noise, making it difficult to establish correspondences on events triggered by the same landmark point. In addition, the asynchronous nature of event data, which is usually obtained from different viewpoints at different timestamps, leads to more complex geometric relationships than those utilized in conventional vSLAM systems. Therefore, new systems for processing the event data need to be developed to fully realize the advantages of event cameras in vSLAM tasks.

In recent years, many event-based vSLAM systems (Fig.~\ref{subfig:event_slam}) have been proposed, which can be categorized into four main types based on their methodologies: feature-based methods, direct methods, motion-compensation methods, and deep learning methods. Feature-based methods extract features from event data \cite{Li2019FAHarrisAF, Rebecq2017RealtimeVO} and establish explicit data association between incoming event data and the extracted features. These methods are able to estimate camera poses and the corresponding 3D feature points simultaneously. Differing from the frame-based direct methods \cite{6906584, 7898369} which construct implicit data association using brightness information, event-based direct methods accumulate event data on the image plane (Fig.~\ref{subfig:event_image}), \textit{e.g.}, edge map \cite{7797445}, and establish data association based on the intensities of the image-alike event representations. Besides, some other event-based direct methods \cite{Kim2016RealTime3R} leverage photometric relationships between brightness changes and absolute brightness intensity to associate events with the corresponding pixels in the reference image. Motion-compensation methods \cite{8578505} recover the motion model by warping and aligning the event data into a reference frame based on their spatio-temporal relationships. Finally, deep learning methods utilize deep neural networks, such as Convolutional Neural Networks (CNNs) and Spiking Neural Networks (SNNs), to process event data and directly predict camera motion and depth.

In this paper, we provide a comprehensive review of event-based vSLAM techniques, as illustrated in Fig. \ref{fig:structure}, covering three primary problems: 1) how to establish event data correspondences explicitly or implicitly, 2) how to obtain accurate camera poses with event data correspondences, and 3) how to build a 3D map from event data. We begin by providing an overview of the general working principle of the event camera, the common event representations, and the general pipeline of event-based vSLAM. Next, we comprehensively review the event-based vSLAM techniques in the aforementioned four main categories and discuss their advantages and disadvantages. We also provide empirical performance evaluation of state-of-the-art (SOTA) event-based vSLAM systems on commonly used datasets and benchmarks. Finally, we discuss the current challenges and suggest some promising directions for future research.

% needed in second column of first page if using \IEEEpubid
%\IEEEpubidadjcol

Several survey papers have been published regarding vSLAM algorithms. For instance, Cadena \textit{et al.} \cite{7747236} propose a standard structure for vSLAM algorithms and discuss various advanced topics in vSLAM such as robustness, scalability, map representation, and theoretical tools. Barros \textit{et al.} \cite{robotics11010024} provide a comprehensive review of vSLAM, visual-inertial SLAM, and RGB-D vSLAM algorithms. However, these papers focus only on vSLAM algorithms that are designed for frame-based cameras and do not include event-based vSLAM algorithms. Recently, there are two surveys \cite{Gallego2019EventBasedVA, Zheng2023DeepLF} related to event cameras. Gallego \textit{et al.} \cite{Gallego2019EventBasedVA} present a review of event cameras, event data processing, and event-based vision tasks. However, they only provide a brief discussion on event-based vSLAM systems and lack many up-to-date references. On the other hand, Zheng \textit{et al.} \cite{Zheng2023DeepLF} emphasize deep learning-based methods in event-based vision tasks. In contrast, we focus on event-based vSLAM algorithms in this work and provide a systematic and comprehensive survey from a wider range of perspectives, including event representation, data correspondence, camera tracking and mapping algorithms, and performance evaluation.

\section{Preliminary}
% -------------------------------------------------------------------------------------------------------------------------------- %
\label{sec:preliminary}

\subsection{Working Principle of Event Camera}
\label{subsec:work_principle}
The working principle of event cameras is fundamentally distinct from conventional frame-based cameras, which capture complete images at a fixed rate, with intensity values for each pixel. On the other hand, event cameras detect brightness changes at each independent pixel and transmit a continuous-time stream of event data, which is asynchronous and timestamped with microsecond resolution. An event can be represented as a tuple $e_k = (x_k, y_k, t_k, p_k)$, where $(x_k, y_k)$ indicates the pixel coordinates that activated the event, $t_k$ represents the timestamp, and $p_k = \pm 1$ indicates the polarity, \textit{i.e.}, the direction of the brightness change.

\textbf{Event Generation Model (EGM).} The event camera sensor operates by first acquiring and storing the logarithmic intensity of brightness, $L(\bm{u}_k) = \log(I(\bm{u}_k))$, at each pixel location $\bm{u}_k$, and then continuously monitoring this intensity value. Whenever the difference in intensity, $\Delta L(\bm{u}_k, t_k)$, between the current and previously recorded values, $L(\bm{u}_k, t_k - \Delta t_k)$, exceeds a specific threshold, $C$, known as the contrast sensitivity, an event $e_k$ is generated by the camera sensor and assigned to the pixel location $\bm{u}_k = (x_k, y_k)$:
\begin{equation}
    \label{eq:egm}
    \Delta L(\bm{u}_k, t_k) = L(\bm{u}_k, t_k) - L(\bm{u}_k, t_k - \Delta t_k) = p_k C,
\end{equation}
where $t_k - \Delta t_k$ is the last recorded timestamp when it triggers an event at the pixel $\bm{u}_k$. Subsequently, the camera sensor stores the current intensity value, $L(\bm{u}_k, t_k)$, and repeats the above process to detect any changes in brightness at this pixel and generate additional events.

\textbf{Linearized EGM.} 
Assuming a constant illumination condition, it is possible to establish a photometric relationship between the change in brightness and the absolute brightness value. Specifically, consider a small time interval, $\Delta t$, during which the change in brightness can be approximated using the information from the images as follows:
\begin{equation}
    \label{eq:linear_egm}
    \Delta L (\bm{u}_k, t_k) \approx - \nabla L(\bm{u}_k) \cdot \bm{v}(\bm{u}_k) \Delta t,
\end{equation}
where $\nabla L(\bm{u}_k) \doteq (\partial_x L, \partial_y L)^{\top}$ is the spatial derivatives of the intensity over $x$ and $y$ coordinates on the image plane and $\bm{v}(\bm{u})$ is the image-point velocity, which is also known as optical flow in the literature. For a more detailed derivation, we refer the reader to the works by Gallego et al. \cite{Gallego2015EventbasedCP, Gallego2019EventBasedVA}. According to the equation, the generation of an event is proportional to the strength of the edges and the amount of motion in the scene. Thus, the edge pattern present in the event data is dependent on the motion of the moving edges. In particular, if the motion occurs parallel to an edge, no events are generated, and the edge becomes invisible in the event data. However, if the motion happens in a different direction from the edge, the moving edge can cause a change in brightness, resulting in the generation of events.

% -------------------------------------------------------------------------------------------------------------------------------- %
\subsection{Event Representation}
% -------------------------------------------------------------------------------------------------------------------------------- %
\label{subsec:event_representation}
Since individual event data carries little information and is subject to noise, events are typically transformed into alternative representations that aggregate meaningful information for vSLAM, to yield a sufficient signal-to-noise ratio. However, event accumulation increases latency, produces motion blur in the representation, and breaks the assumption about events triggered at the same camera location. This section reviews the typical event representations used in event-based vSLAM methods, which include individual event and accumulation representations such as event frame, time surface, and voxel grid.

\textbf{Individual Event.}
Individual event can be directly used in filter-based models, such as probabilistic filters \cite{Kim2016RealTime3R} and SNNs \cite{DBLP:conf/icra/GehrigSMS20}, on a per-event basis. They update internal states asynchronously with each incoming event by reusing the states from past events or external sources (\textit{e.g.}, IMU data).

\textbf{Event Packet.} 
Event packet is another type of event representation that stores a sequence of event data directly in a temporal window. Similar to the individual event, event packets retain precise information such as timestamps and polarities. Since event packets aggregate event data within temporal windows, it allows for batch operations in filter-based methods \cite{6906931} and facilitates the search for optimal solutions in optimization methods \cite{8885542, 7758089}.

\textbf{Event Frame.} 
Event frame is a 2D representation that accumulates event information at the same pixel position, making it a compact event representation. It can be obtained by converting a stream of events into an image-like representation, and then used as input for conventional frame-based vSLAM algorithms, under the assumption that the pixel positions remain constant. For instance, EVO \cite{7797445} transforms event data into an edge map to extract spatial edge patterns from a short temporal window of events. \cite{9341224} proposes event counting to generate a 2D histogram of the number of events for each polarity.

\textbf{Time Surface.}
Time surface (TS) is a 2D representation in which each pixel records a single time value, typically the most recent timestamp of the event that occurred at that pixel. To prevent the timestamps from going to infinity, a normalization process is applied to rescale timestamps to a range of $[0, 1]$, while preserving the distribution of the timestamps. The exponential decay kernel \cite{7508476} is a typical normalization model used in vSLAM algorithms to emphasize recent events over past events.

\textbf{Motion-compensated Event Frame.}
Similar to frame-based cameras, motion blur can also occur in accumulation-based representations of event data. To address this issue, a motion-compensated event frame, known as an image of warped events (IWE), can be used. It warps the event data to a reference frame using a specified motion model, ensuring the correct spatial edge patterns over a long temporal window. For instance, visual-inertial odometry (VIO) methods \cite{Rebecq2017RealtimeVO, Guan2022PLEVIORM} leverage external information such as IMU data to estimate camera motion and reproject the event data to reference timestamps accordingly.

\textbf{Voxel Grid.} 
Voxel grid encodes event data into a fixed-size 3D space-time tensor representation, in which each voxel denotes a particular pixel and a discrete time interval. \cite{MostafaviIsfahani2021LearningTR} presents a simple and straightforward event stacking method, which merges and stacks event data in a voxel by summing the polarities. In order to improve temporal resolution beyond the number of bins, the discretized event volume (DEV) \cite{8953979} utilizes a linearly weighted accumulation of the polarities of all neighborhoods at each voxel.

\textbf{Reconstructed Image.}
Reconstructed images usually refer to brightness images that are obtained by merging an initial brightness image with incremental changes in brightness provided by the event data. For instance, deep learning approaches \cite{8946715, MostafaviIsfahani2021LearningTR} convert raw event data to voxel grids and reconstruct the brightness images via supervised training. While the reconstructed images can restore the brightness information in HDR conditions and reduce motion blur, they may lack essential textures for vSLAM algorithms since event data only responds to the motion of scene edges.

\subsection{General Pipeline of Event-based vSLAM}
\label{subsec:pipeline}
Most event-based vSLAM systems follow the pipeline of the PTAM \cite{4538852}, which consists of three main components: 1) initialization, 2) camera tracking, and 3) mapping. In initialization, an initial map and camera poses are roughly estimated, after which camera tracking and mapping iteratively estimate the camera poses and reconstruct the 3D scene map. Alternatively, other approaches \cite{9811943} perform the tracking and mapping steps simultaneously by optimizing the camera poses and 3D feature positions jointly. Parallel optimization is advantageous in terms of computational efficiency, whereas joint optimization helps to prevent the accumulation of drift errors in the two threads.

\textbf{Tracking.}
The camera tracking module estimates the camera poses with respect to the current reconstructed 3D map. The camera tracking module first constructs the 2D-3D data correspondences explicitly or implicitly. Then, the camera poses that best bit the data correspondences are estimated by optimizing the reprojection errors between 2D pixels and reprojected 3D points.

\textbf{Mapping.}
The mapping module aims to reconstruct the 3D structure of the environment. The corresponding 2D pixels from different camera views are triangulated to estimate the 3D point. Usually, the depth value from the reference frame is estimated, which reduces the dimension of the 3D point space. Moreover, since the size of the object occupies one pixel is relative to the distance from the object to the camera, the inverse depth \cite{7797445} and log depth representation \cite{9320311} are often chosen to achieve high-granularity with small depth and low-granularity with large depth.

\textbf{Mapping Representation.}
Due to the sparsity property of event data, event-based vSLAM algorithms usually reconstruct semi-dense structures using point-based and line-based map representations, including voxel grid, depth map and 3D lines.Voxel grid \cite{6906882, 10.1007/s11263-017-1050-6} discretizes the 3D volume into a finite set of voxels, which represents the existence of 3D points. Depth map estimates the depth pixelwise in the reference frame, whose value represents the depth in the corresponding pixel from the reference camera view in \cite{7758089, Kim2016RealTime3R}. Contrast to the point-based map representation, 3D lines map \cite{Gentil2020IDOLAF, 9810191} represents the edges which the event camera naturally responds to.

\textbf{Map Density. }
The mapping module will finally reconstruct a sparse, semi-dense or dense 3D map. Generally, event-based vSLAM methods consider the 3D map from all the event data as a semi-dense scene map \cite{7797445}, while a few methods \cite{7758089, Guan2022PLEVIORM} first selects a subset of events to effectively construct a 3D sparse map for the time efficiency and accuracy. Since the event data do not contain full information from 2D view of the 3D scene, the event-based vSLAM algorithms are usually unable to produce the dense 3D map. To handle this issue, deep learning-based methods leverage the generalized ability of deep learning models to recover the dense 3D scene map.

% -------------------------------------------------------------------------------------------------------------------------------- %
\section{Feature-based Method}
% -------------------------------------------------------------------------------------------------------------------------------- %
\label{sec:feature}

\begin{table*}
    \centering
    \caption{A summary of existing event-based vSLAM methods. (Opt: Optimization, DL: Deep Learning, MC: Motion-compensated, Rot: Rotational, E: Event Camera, F: Frame-based Camera, D: Depth Sensor, I: Inertial Sensor, SE: Stereo Event Camera)}
    \resizebox{2.03\columnwidth}{!}{
    \begin{tabular}{l | c | c | c | c | c | c | c | l }
    \hline
         Reference & Method & Type & Event Representation & Motion & Scene & Sensor & Real-Time & Remarks \\
        \hline
        Cook \textit{et al.} \cite{6033299} & Opt & - & Event Frame & Rot & Natural & E & \xmark & Interacting network \\
        Weikersdorfer \textit{et al.} \cite{6491077} & Filter & Feature & Individual Event & Planar & 2D B\&W & E & \cmark & First filter-based VO \\
        Weikersdorfer \textit{et al.} \cite{10.1007/978-3-642-39402-7_14} & Filter & Feature & Individual Event & Planar & 2D B\&W & E & \cmark & Planar probabilistic map \\
        M$\mathrm{\ddot{u}}$ggler \textit{et al.}\cite{6942940} & Opt & Feature & Event Packet & 6DoF & 2D B\&W & E & - & Line-based VO on known patterns \\
        Kim \textit{et al.} \cite{BMVC.28.26} & Filter & Direct & Individual Event & Rot & Natural & E & \cmark & Two interleaved filters \\
        Censi \textit{et al.} \cite{6906931} & Filter & Direct & Event Packet & 6DoF & B\&W & E+F+D & - & Filter-based VO based on image gradient \\
        Weikersdorfer \textit{et al.} \cite{6906882} & Filter & Feature & Individual Event & 6DoF & Natural & E+D & \cmark & Augment with depth sensor \\
        Gallego \textit{et al.} \cite{Gallego2015EventbasedCP} & Filter & Direct & Individual Event & 6DoF & Natural & E & - & Asynchronous event-image alignment \\
        Mueggler \textit{et al.} \cite{Mueggler2015ContinuousTimeTE} & Opt & Feature & Event Packet & 6DoF & 2D B\&W & E & \xmark & Continuous camera trajectory \\
        Yuan \textit{et al.} \cite{7487657} & Opt & Feature & Event Frame & 6DoF & Natural & E+I & - & Vertical line-based VO\\
        Kueng \textit{et al.} \cite{7758089} & Opt & Feature & Local Point Set & 6DoF & Natural & E+F & \xmark & Event-based feature tracking VO \\
        Kim \textit{et al.} \cite{Kim2016RealTime3R} & Filter & Direct & Individual Event & 6DoF & Natural & E & \cmark & Three interleaved fitlers \\
        EVO \cite{7797445} & Opt & Direct & Edge Map & 6DoF & Natural & E & \cmark & Event-event geometric alignment \\
        Gallego \textit{et al.} \cite{8094962} & Filter & Direct & Individual Event & 6DoF & Natural & E+F+D & \xmark & Resilient sensor model \\
        CMax \cite{7805257, 8578505} & Opt & Motion & MC Event Frame & Rot & Natural & E & \cmark & Contrast maximization \\
        Reinbacher \textit{et al.} \cite{DBLP:journals/corr/ReinbacherMP17} & Opt & Feature & Event Packet & Rot & Natural & E & \cmark & Panoramic probabilistic map \\
        Bryner \textit{et al.} \cite{8794255} & Opt & Direct & Event Frame & 6DoF & Natural & E & \xmark & Synchronous event-image alignment \\
        Zhu \textit{et al.} \cite{8961878} & Opt & Feature & MC Event Frame & 6DoF & Natural & E & \xmark & Probabilistic feature tracking VO \\
        Zhu \textit{et al.} \cite{8953979} & Opt & DL & Voxel Grid & 6DoF & Natural & E & - & Voxel grid input and MC loss function \\
        Ye \textit{et al.} \cite{9341224} & Opt & DL & Event Frame, Time Surface & 6DoF & Natural & E & \cmark & Event-based SfMLearner \\
        Xu \textit{et al.} \cite{8950341} & Opt & Motion & MC Event Frame & Rot & Natural & E & \xmark & Smooth Constraint \\
        DMin \cite{10.1007/978-3-030-58558-7_10, 9625712} & Opt & Motion & MC Event Features & 6DoF & Natural & E+D & \cmark & Dispersion minimization \\ 
        Liu \textit{et al.} \cite{9156854} & Opt & Motion & MC Event Frame & Rotational & Natural & E & \xmark & Globally optimal CMax \\
        Peng \textit{et al.} \cite{10.1007/978-3-030-58574-7_4, 9329204} & Opt & Motion & MC Event Frame & Planar, Rot & Natural & E & \xmark & Globally optimal CMax \\
        Bertrand \textit{et al.} \cite{9291346} & Opt & Feature & Individual Event & 6DoF & 2D B\&W & E & \cmark & Embedded VO system \\
        Chamorro \textit{et al.} \cite{Chamorro:BMVC20, 9810191} & Filter & Feature & Individual Event & 6DoF & Natural & E & \cmark & Line-based VO \\
        Gehrig \textit{et al.} \cite{DBLP:conf/icra/GehrigSMS20} & Opt & DL & Individual Event & Rot & Natural & E & \xmark & Shallow SNN \\
        ST-PPP \cite{Gu_2021_ICCV} & Opt & Motion & MC Event Frame & Rot, Planar & Natural & E & \cmark & MC probabilistic model \\
        Kim \textit{et al.} \cite{9454404} & Opt & Motion & MC Event Frame & Rot & Natural & E & \cmark & Global MC event alignment \\
        Liu \textit{et al.} \cite{Liu2021SpatiotemporalRF} & Opt & Feature & Event Packet & Rot & Natural & E & - & Spatio-temporal consistency \\
        VCM \cite{s22155687} & Opt & Motion & Event Packet & Planar & Natural & E & \cmark & CMax in 3D space \\
        Zuo \textit{et al.} \cite{9811805} & Opt & Direct & Time Surface & 6DoF & Natural & E+D & \cmark & Augment depth sensor in mapping \\
        Liu \textit{et al.} \cite{9811943} & Opt & Feature & Individual Event & 6DoF & Natural & E & \xmark & Continuous-time incremental SfM \\
        EDS \cite{9879881} & Opt & Direct & Event Frame & 6DoF & Natural & E+F & \xmark & Synchronous event-image alignment \\
        ESVO \cite{9386209} & Opt & Direct & Time Surface & 6DoF & Natural & SE & \cmark & First stereo VO \\
        Hadviger \textit{et al.} \cite{Hadviger2021FeaturebasedES} & Opt & Feature & Time Surface & 6DoF & Natural & SE & \cmark & Stereo feature matching VO \\
        Mueggler \textit{et al.} \cite{8432102} & Opt & Feature & Event Packet & 6DoF & Natural & E+I & \xmark & Continuous-time camera trajectory\\
        Zhu \textit{et al.} \cite{8100099} & Filter & Feature & MC Event Packet & 6DoF & Natural & E+I & \xmark & Probabilistic feature association VIO \\
        Rebecq \textit{et al.} \cite{Rebecq2017RealtimeVO} & Opt & Feature & MC Event Frame & 6DoF & Natural & E+I & \cmark & Joint optimization with IMU data\\
        USLAM \cite{Vidal2017UltimateSC} & Opt & Feature & MC Event Frame & 6DoF & Natural & E+F+I & \cmark & Feature tracking on events and images \\ 
        IDOL \cite{Gentil2020IDOLAF} & Opt & Feature & Event Packet & 6DoF & Natural & E+I & \xmark & Gaussian pre-integrated measurement \\
        EIO \cite{9981970} & Opt & Feature & Time Surface & 6DoF & Natural & E+I & \cmark & First VO with loop closure \\
        PL-EVIO \cite{Guan2022PLEVIORM} & Opt & Feature & MC Time Surface & 6DoF & Natural & E+F+I & \cmark & Point and line features tracking VIO \\
        Mahlknecht \textit{et al.}  \cite{Mahlknecht2022ExploringEC} & Filter & Feature & Event Frame & 6DoF & Natural & E+F+I & \xmark & Robust event-based xVIO \\
        ESVIO \cite{s23041998} & Opt & Direct & Time Surface & 6DoF & Natural & SE+I & \cmark & Integrate stereo VO with IMU data \\
        \hline
    \end{tabular}
    }
\end{table*}

\begin{figure}
    \centering
    \resizebox{0.75\columnwidth}{!}{
    \begin{tikzpicture}[node distance=1cm]
        \node [process, anchor=north] (event) {Event Data};
        \node [process, below of=event, yshift=-0.2cm, text width=3.5cm] (e-rep) {Event Representation};
        \draw [flow line] (event) -- (e-rep);

        \node [process, below of=e-rep, xshift=-2.5cm, yshift=-0.4cm] (f-e) {Feature Extraction};
        \node [process, below of=e-rep, xshift=2.5cm, yshift=-0.4cm] (f-t) {Feature Tracking};
        \draw [flow line] (e-rep) -- (f-e);
        \draw [flow line] (e-rep) -- (f-t);
        \draw [flow line] (f-e) -- node[above]{features} (f-t);

        \node [process, below of=e-rep, xshift=-2.5cm, yshift=-3.0cm] (pose) {Camera Tracking};
        \node [process, below of=e-rep, xshift=2.5cm, yshift=-3.0cm] (map) {3D Mapping};
        \node at ($(pose.north)!.5!(map.north)+(0,0.3)$) [] (p) {Pose};
        \node at ($(pose.south)!.5!(map.south)-(0,0.3)$) [] (d) {Depth};
        \draw [flow line] ($(pose.north)+(0.3,0.0)$) |- (p) -| ($(map.north)-(0.3,0.0)$);
        \draw [flow line] ($(map.south)-(0.3,0.0)$) |- (d) -| ($(pose.south)+(0.3,0.0)$);
        \begin{scope}[on background layer]
            \node [box, draw=black, fit=(pose)(map)(p)(d)] (tam) {};
        \end{scope}
        \draw [flow line] (f-t) |- ($(tam.north)!.5!(f-t.south)$) node[above,xshift=-1.0cm]{feature correspondences} -| (tam);
        \node [below of=pose, yshift=-0.6cm] (out_p) {Pose};
        \node [below of=map, yshift=-0.6cm] (out_d) {3D Map};
        \draw [flow line] (pose) -- (out_p);
        \draw [flow line] (map) -- (out_d);
    \end{tikzpicture}
    }
    \caption{The flow diagram describes the process of the feature-based visual odometry (VO) algorithms with pure event data.}
    \label{fig:feature_method}
\end{figure}
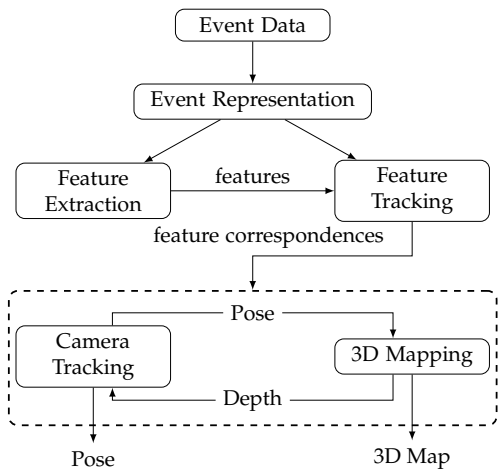

As shown in Fig. \ref{fig:feature_method}, feature-based vSLAM algorithms consist of two primary components: 1) feature extraction and tracking, and 2) camera tracking and mapping. In the feature extraction step, robust features that are invariant to various factors, including motion, noise, and illumination changes, are identified. The subsequent feature tracking step is then utilized to associate features corresponding to the same scene points. Using the associated features, camera tracking and mapping algorithms estimate the relative camera poses and 3D landmarks of the features simultaneously. In this section, we review relevant approaches in the event domain.

\subsection{Feature Extraction}
\label{subsec:feature_extraction}
Feature extraction methods detect shape primitives in the event stream, including point-based and line-based features. Point-based features represent specific points of interest, \textit{e.g.}, the intersection of event edges, while line-based features are clusters of events lying on the same straight lines.

\subsubsection{Point-based Features}
Point-based features, also known as corners, are defined as the intersection of two or more event edges on the image plane. In order to detect corners within the event data, \cite{CLADY201591} and \cite{10.3389/fnins.2016.00594} employ the local plane fitting algorithm to detect the normal planes of the optical flow and label an event as a corner if it belongs to the intersection of two or more of these planes. However, this technique suffers from high computational complexity due to the multiple optimization steps involved in feature computation.

One straightforward approach is to apply frame-based corner detectors to the event representation, such as TS. For example, the eHarris method \cite{7759610} binarizes TS and applies the derivative filter of the Harris detector \cite{Harris1988ACC} to classify corners. The continuous Harris event corner detector (CHEC) \cite{8613800} proposes a local spatial convolution operation on a sequence of events to estimate image gradients with the Sobel kernel and update the Harris score incrementally for corner classification. The eFAST corner detector \cite{Mueggler2017FastEC} utilizes the FAST detector \cite{Rosten2006MachineLF} to improve computational efficiency. It detects corners by searching for contiguous pixels with higher values than the rest on two concentric circles around the current event. To avoid misclassifying corners on edges, the maximum angle of the circular arc is restricted to a value less than $180^{\circ}$ in this approach. To address this issue, the Arc$*$ algorithm \cite{8392795} iteratively expands circular arcs from the latest event in clockwise and counter-clockwise directions for corner classification. However, the detection of obtuse angles results in many false corner detections along edges. To reduce the number of false detections, the FA-Harris method \cite{Li2019FAHarrisAF} uses the eHarris detector to filter out corner candidates obtained from the Arc$*$ algorithm.

The Harris and FAST detectors have also been extended to different types of event representations, including the edge map \cite{8100099, 7989517} and motion-compensated event frame \cite{Rebecq2017RealtimeVO, Vidal2017UltimateSC}. Specifically, \cite{7989517} applies the Harris detector on the edge map, while \cite{8100099} utilizes the FAST detector to extract corner candidates and selects the candidate with the highest Shi-Tomasi score within each cell as a corner. Additionally, \cite{Rebecq2017RealtimeVO} and \cite{Vidal2017UltimateSC} construct motion-compensated event frames over a longer temporal window of event data to eliminate motion blur and then detect corners using the FAST detector.

The aforementioned hand-crafted feature extraction methods suffer from the sensitivity to motion changes of features and noise in event cameras. To overcome this limitation, learning-based methods \cite{8954376, 9522732} have been proposed to model motion-variant corner patterns and event noises for improved corner detection stability. \cite{8954376} introduces the speed-invariant time surface and applies a random forest to predict corners. To alleviate the need for feature annotation, \cite{9522732} trains a ConvLSTM \cite{Shi2015ConvolutionalLN} to predict image gradients from event data, with supervision from brightness images, and applies the Harris detector on the predicted gradients.

\subsubsection{Line-based Features}
Since event cameras inherently respond to moving edges and the environment usually contains a significant number of lines, several algorithms have been proposed to extract line-based features from event data. Among these, some works use classical line detection algorithms such as the Hough transformation and line segment detector (LSD) \cite{4731268}. Other approaches exploit the spatio-temporal relationship in event data \cite{10.3389/fnbot.2018.00004} or leverage external IMU data to cluster events in vertical bins \cite{7487657}.

For example, \cite{6942940} utilizes the Hough transformation for detecting lines in the event stream by projecting events onto the Hough space. To use a flexible threshold and leverage the neighbors' information, the spiking Hough transformation algorithm is proposed in \cite{9291346}, which employs spiking neurons in the Hough space to detect line. Furthermore, \cite{9810191} extends the Hough transformation to a 3D point-based map to group event data to a set of 3D lines.

PL-VIO \cite{Guan2022PLEVIORM} performs line-based feature extraction by directly applying the LSD algorithm on the motion-compensated event stream.
Inspired by the LSD algorithm, the Event-based Line Segment Detector (ELiSeD) \cite{7605244} computes the orientation of each event in TS by utilizing the Sobel filter and groups events with similar angles into a line support region.
Similarly, \cite{8463622} computes the orientation of event data based on optical flow and assigns events with similar distances and orientations to a line cluster.

Based on the observation that a moving line with constant velocity triggers a set of events on a plane in the spatio-temporal space, the work \cite{10.3389/fnbot.2018.00004} applies a plane fitting algorithm to cluster events and estimate the line as the cross product of the optical flow and the plane normal. In another work, \cite{7487657} aligns event data to the gravity direction of the world, and uses vertical bins to cluster event data, thus enabling the extraction of vertical lines.

\subsection{Feature Tracking}
\label{subsec:feature_tracking}
Given a set of features, feature tracking algorithms aim to associate each event with its corresponding features.
Feature tracking algorithms usually update the parametric models of features templates, such as 2D rigid body transformation, feature motion trajectories and feature positions.
Other methods leverage the descriptor matching to establish feature correspondences while deep learning methods apply DNNs to predict feature displacements.

The parametric transformation, such as the Euclidean transformation, is commonly employed to model the positions and orientations of event-based features on the image plane for feature tracking and correspondence. Feature patterns can be constructed from local patches of either predefined features in \cite{6204348, 7063371}, or Canny edge maps around the Harris corners from the image in \cite{7605086, 7758089}. The extracted features are then tracked with subsequent events using the iterative closest point (ICP) algorithm \cite{121791} under the registration transformation in \cite{6204348, 7063371, 7605086, 7758089}. However, neither \cite{7605086} nor \cite{7758089} account for edges strength, where all edges are indistinguishable in the binary edge map. In contrast, the EKLT tracker \cite{Gehrig2018EKLTAP} estimates brightness changes as feature patterns based on the linearized EGM, where image gradients are utilized to measure the edge strength.

Feature tracking and correspondence establishment typically require modeling feature motions on the image plane. Two expectation-maximization optimization steps are introduced in \cite{7989517, 8100099} to estimate the optical flow of events and the affine transformation of feature alignment. The Lucas-Kanade optical flow tracker is applied to the motion-compensated event frame in \cite{Vidal2017UltimateSC} and the TS in \cite{Guan2022PLEVIORM} to track features. It should be noted that the use of optical flow implicitly assumes a linear model for feature trajectories, which can be easily violated in practice. To address this issue, continuous curve representations such as B$\acute{e}$izer curves \cite{Seok_2020_WACV} and B-splines \cite{Chui2021EventBasedFT} have been explored for feature trajectory modeling. Contrast maximization (CMax) methods can then be used to maximize the event alignment along the curves for feature trajectory estimation.

Tracking and associating features based on their discrete positions on the image plane has also been investigated. In \cite{8392795} and \cite{8491018}, a graph is constructed with nodes representing event features and edges representing feature associations for feature tracking. Proximity \cite{8392795} and local region descriptors \cite{8491018} are used to match new event data with corresponding graph nodes. However, tracking features based on their positions on a per-event basis can make the algorithms sensitive to event noise. To address this issue, \cite{8885542} and \cite{alzugaray:BMVC20} propose to discretize spatial neighborhoods of each feature into a set of candidate feature points, and updates alignment scores between feature templates and multiple candidates. A candidate becomes a new corresponding feature, if it outperforms the previous feature. This process is repeated to continuously track the features. In \cite{9981451}, events with equal polarity are grouped as a set of equal polarity cluster features based on the observation that feature patterns of different polarities create a boundary separating clusters of events. After the feature clustering, the feature tracks are estimated as the center of gravity of events in a set of temporal windows.

Instead of modeling feature motion on the image plane, feature descriptors can be used to establish feature correspondences directly. For instance, \cite{Hadviger2021FeaturebasedES} builds the feature descriptor as a square window centered at the event corner in TS and establishes correspondences via cross-correlation between the feature descriptors. Meanwhile, PL-EVIO \cite{Guan2022PLEVIORM} utilizes lines as features and applies the band descriptor \cite{Zhang2013AnEA} to describe and match line-based features.

Traditional feature tracking methods based on linear noise models cannot handle the non-linear noise characteristics of event cameras. In contrast, deep learning methods can implicitly model event noises and generalize well across different scenarios under the assumption that the training and test datasets exhibit only minor distribution shifts. For example, \cite{Messikommer2022DatadrivenFT} proposes a deep learning-based approach that extracts local contextual information and correlation maps for each feature and the corresponding local patch of event data with the feature pyramid network (FPN) \cite{Lin2016FeaturePN} and ConvLSTM \cite{Shi2015ConvolutionalLN}, and predicts the feature displacement accordingly. Moreover, it incorporates a frame attention module to aggregate the information from all features and refine the predictions. To narrow down the distribution shift, the network is trained on synthetic data and finetuned with additional pose supervision on real data.

\subsection{Camera Tracking and Mapping}
\label{subsec:feature_tam}

Given the event data associations, most features-based vSLAM algorithms perform the tracking and mapping tasks in parallel threads. In the tracking thread, the feature-based techniques estimate the relative camera poses using a set of 3D feature positions obtained from the mapping thread, while these feature positions are used to reconstruct the 3D landmarks of features in the mapping thread in return. In this section, we present a comprehensive review of four typical types of event-based vSLAM approaches based on: 1) the conventional frame-based vSLAM methods, 2) filter-based methods, 3) continuous-time camera trajectory methods, and 4) spatio-temporal consistency methods.

\textbf{Frame-based Method.}
Based on the 2D image-like event representation, conventional frame-based vSLAM algorithms can be adapted for event-based tracking and mapping. For instance, \cite{6942940} and \cite{9291346} apply the reprojection error between the reprojected line and the associated event data, and between the intersection points of lines and the corresponding 3D points, respectively, to estimate camera poses with respect to the known planar shape. \cite{7758089} and \cite{8961878} utilize the SVO algorithm \cite{6906584} to estimate the relative camera pose by minimizing the reprojection error on a set of event-feature correspondences. In the mapping module, these methods apply depth filters to measure the depth values as a mixture of Gaussian and uniform distribution and update depth values as well as uncertainties by the features triangulation between the current location and its first detection, using the relative camera poses. Additionally, \cite{7487657} derives a linear system to represent the algebraic distance of 2D to 3D vertical lines and solve the camera poses using least squares optimization.

\textbf{Filter-based Method.}
Frame-based vSLAM methods assume temporally synchronized 2D frames, ignoring the asynchronous nature of event data. Filter-based methods have been proposed to handle event data asynchronously. Typically, the filter's state is defined as the current camera pose, while the motion model is the random diffusion model. The state is predicted using the motion model and then corrected using error measurement. For instance, \cite{6491077} measures the distance between the back-projected ray of an event data and a 3D planar feature point to estimate the camera pose. \cite{10.1007/978-3-642-39402-7_14} extends this method for mapping with an additional measurement function based on the probability of event occurrence in the planar plane. Furthermore, line-based vSLAM methods \cite{Chamorro:BMVC20, 9810191} update the filter state during camera tracking by measuring the distance between an event and the reprojected line. During mapping, line-based methods use Event-Based Multi-View Stereo (EMVS) \cite{10.1007/s11263-017-1050-6} to construct a point-based map and apply the Hough transformation to extract 3D lines, which implicitly indicate event-line associations.

\textbf{Continuous-time Method.}
Filter-based methods, while capable of handling asynchronous data, often require a large number of parameters (\textit{i.e.}, control states) for camera pose representation. To address this issue, continuous-time representation of camera trajectory has been introduced as a promising solution. By representing the camera trajectory as a continuous curve, such as B-spline in \cite{Mueggler2015ContinuousTimeTE, 8432102} or Gaussian process motion model in \cite{9811943, Gentil2020IDOLAF}, these methods can interpolate the camera pose at any given time from the local control states and use it to obtain optimal control states. Notably, \cite{9811943} proposes a joint optimization method based on incremental Structure from Motion (SfM) to simultaneously update the control states and 3D landmarks.

\textbf{Spatio-Temporal Consistency Method.}
By assuming any two events with the same time interval following the same relative rotational transformation, \cite{Liu2021SpatiotemporalRF} derives a spatio-temporal consistency constraint for events under rotational camera motion. To optimize the motion parameters, it iteratively searches for the closest points based on a relaxed temporal constraint and applies the trimmed ICP algorithm to enforce spatial consistency.

\subsection{Multi-sensor Method}

\textbf{RGB-D Sensor.}
\cite{6906882} extends \cite{10.1007/978-3-642-39402-7_14} by incorporating depth information from a depth sensor to enabling 3D probabilistic map reconstruction, which requires additional event and depth correspondences.

\textbf{IMU Sensor.}
Since events are motion-dependent, events do not contain long-term appearance, which leads to unreliable feature extraction algorithm. To address this issue, \cite{Rebecq2017RealtimeVO} and \cite{ Vidal2017UltimateSC} aggregate events over a long temporal window into motion-compensated event images to preserve the long-term appearance and apply frame-based feature extraction algorithms to acquire reliable features. On the other hand, several event-based VIO methods \cite{Rebecq2017RealtimeVO, Vidal2017UltimateSC, 9981970, Guan2022PLEVIORM} employ the IMU pre-integration algorithm \cite{leutenegger2013keyframe, 7557075} to incorporate event data with inertial data and apply the sliding-window optimization methods to minimize the IMU error and the reprojection error of event features. Furthermore, filter-based VIO methods \cite{8100099, Mahlknecht2022ExploringEC} improve the motion model by propagating the last IMU state to the event state at the corresponding timestamp, which avoids failures in camera tracking and improves the accuracy. Moreover, the continuous-time representation \cite{8432102, Gentil2020IDOLAF} can query the camera state at any timestamp, which enables the fusion of asynchronous event data and synchronous IMU data.

\textbf{Image Sensor.}
Since the images contain abundant static long-term features and events are of high temporal resolution, the works \cite{7605086, 7758089, Gehrig2018EKLTAP} propose to extract features (\textit{e.g.} the Canny edge map around Harris corner) from the last image and track these features using event data. Similarly, \cite{Messikommer2022DatadrivenFT} inputs both images and events to extract their local contextual information, and constructs correspondences between events and image features. On the other hand, \cite{9522732} utilizes the image gradients from images as supervised signals and trains a ConvLSTM \cite{Shi2015ConvolutionalLN} with event as input only to predict image gradients. Afterwards, the Harris corner detector is applied on the predicted gradients.

\subsection{Discussion}
The accuracy of feature extraction and tracking algorithms has a significant impact on the performance of feature-based methods. Current feature extraction algorithms rely heavily on frame-based feature detection, such as Harris and Fast corner detector. However, event data poses unique challenges as its appearance is dependent on the motion, unlike intensity images. As a result, the constant appearance assumption underlying frame-based algorithms is not applicable to event data. To address this issue, some works \cite{7758089, Gehrig2018EKLTAP} have proposed extracting motion-invariant features from images and tracking them using event streams. Nonetheless, these methods suffer from limitations of frame-based cameras such as motion blur and low dynamic range. Learning-based methods are capable of efficiently detecting stable feature points using pure event data. However, most learning-based methods require intensity images to provide supervised signals for model training. Feature-based VIO methods, on the other hand, couple IMU data with event data to create motion-compensated event frames over a long temporal window. The long-term appearance in the motion-compensated event frames is less affected by noise and motion-variant characteristics, enabling robust feature extraction and tracking. However, this approach is deprived of the high temporal resolution of event data.

% -------------------------------------------------------------------------------------------------------------------------------- %
\section{Direct Method}
% -------------------------------------------------------------------------------------------------------------------------------- %
\label{sec:direct}

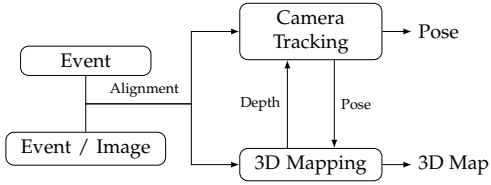
\begin{figure}
    \centering
    \resizebox{0.75\columnwidth}{!}{
        \begin{tikzpicture}[node distance=1cm]
            \node [process, anchor=west, text width=2.0cm] (event) {Event};
            \node [process, below of=event, yshift=-0.5cm, text width=2.5cm] (image) {Event / Image};

            \node [process, right = 1.5cm of event, yshift=.5cm] (track) {Camera Tracking};
            \node [process, below = 1.5cm of track] (map) {3D Mapping};
            \draw [flow line] (event) |- node[anchor=south ,xshift=1.0cm,font=\scriptsize]{Alignment} ($(event.south)!.5!(image.north)+(1.8,0.0)$) |- (track) ;
            \draw [flow line] (image) |- ($(event.south)!.5!(image.north)+(1.8,0.0)$) |- (map) ;
            \draw [flow line] ($(track.south)+(0.4,0.0)$) -- node[anchor=west ,xshift=0.0cm,font=\scriptsize]{Pose} ($(map.north)+(0.4,0.0)$) ;
            \draw [flow line] ($(map.north)-(0.4,0.0)$) -- node[anchor=east ,xshift=0.0cm,font=\scriptsize]{Depth} ($(track.south)-(0.4,0.0)$) ;

            \node [right = 0.5cm of track] (pose_2) {Pose};
            \node [right = 0.5cm of map] (depth_2) {3D Map};
            \draw [flow line] (track) -- (pose_2);
            \draw [flow line] (map) -- (depth_2);

        \end{tikzpicture}
    }
    \caption{
    The diagrams depict event-based direct methods. Direct methods attempt align events data to the corresponding events or pixels in image to estimate camera poses and 3D maps.
    }
\end{figure}
Unlike feature-based methods, direct methods align event data in camera tracking and mapping algorithms without explicit data association. Frame-based direct methods estimate relative camera poses and 3D positions by comparing pixel intensities between selected pixels in a source image and the corresponding reprojected pixels in the target image. However, they cannot be applied to event streams due to the asynchronous nature and brightness change information of event data. To overcome this challenge, two types of event-based direct methods have been developed: event-image alignment and event representation-based alignment. In this section, we provide a comprehensive review of these two categories of event-based direct methods.

\subsection{Event-Image Alignment Method}
\label{subsec:event_image_align}

\begin{figure}
    \centering
    \includegraphics[width=0.35\textwidth]{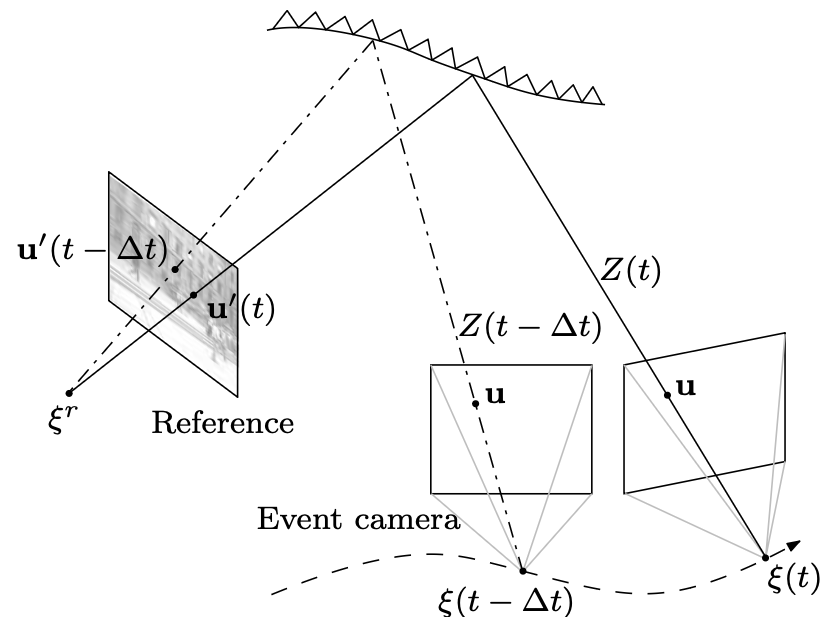}
    \caption{Computation of the contrast threshold by reprojecting events from the event camera to a reference image. Figure adapted from \cite{8094962}.}
    \label{fig:egm}
\end{figure}

Event-image alignment methods utilize additional visual images to ensure photometric consistency between event data and images. Based on the event generation mechanism (Eq. \ref{eq:egm}), whereby each event represents the brightness change from the last event triggered at the same pixel, several works \cite{BMVC.28.26, Kim2016RealTime3R, 8094962} align event data with the corresponding brightness pixels to estimate camera poses and depths, as shown in Fig. \ref{fig:egm}. In direct methods, filter-based techniques are also employed to process incoming event data. For instance, \cite{BMVC.28.26} utilizes two filters to estimate camera poses and image gradients for image reconstruction under rotational camera motion. In the camera tracking module, the first filter utilizes EGM by re-projecting events to a reference image to calculate the brightness difference as the error measurement to update the camera poses. The second filter estimates the image gradient at each pixel on the reference images using the linearized EGM as the measurement model to recover the absolute brightness intensity. Furthermore, \cite{Kim2016RealTime3R} uses an extra filter for depth estimation, which applies the same error measurement function in the camera tracking module to update the depth values in the reference frame. To enhance the robustness of filter-based methods, \cite{8094962} utilizes a resilient sensor model with a normal-uniform mixture distribution to filter out outliers in event data.

Several approaches have been proposed to estimate camera poses and velocities using event data. \cite{6906931} employs a filter to maximize the magnitude of camera velocity and image gradients, as events are more likely to occur in regions with large brightness gradients. Similarly, \cite{Gallego2015EventbasedCP} uses the linearized EGM to obtain the camera motion parameters by constructing an implicit measurement function. Although previous methods update camera poses on a per-event basis using filtering, this can be computationally expensive. To address this issue, \cite{8794255} and \cite{9879881} process a group of events synchronously using non-linear optimization. They convert an event stream to a brightness incremental image and align it with a reference brightness incremental image to estimate camera pose and velocity simultaneously. In the mapping module, \cite{Gallego2015EventbasedCP, 8794255} assume a given photometric 3D map consisting of intensities and depths, while \cite{9879881} transfers depth values from past keyframes to a new keyframe and applies photometric bundle adjustment (PBA) on keyframes to refine the camera poses and 3D structure.

\subsection{Event Representation-based Alignment Method}
\label{subsec:event_event_align}
Event-image alignment methods require additional data, such as a photometric 3D map consisting of intensities and depths, as well as brightness images, to ensure photometric consistency. Conversely, event representation-based alignment methods follow the same scheme of the frame-based direct method \cite{6906584}, where event data is aligned by converting it into 2D image-like representations. EVO \cite{7797445} presents a purely geometric method for aligning event data based on edge patterns. In the camera tracking module, it converts a sequence of events to an edge map, which is then aligned with the reference frame built from the reprojection of the 3D map. The mapping module adopts EMVS \cite{10.1007/s11263-017-1050-6} to reconstruct a local semi-dense 3D map without explicit data associations. ESVO \cite{9386209} presents a direct method on TS to exploit temporal information from event data. In the camera tracking module, it aligns the support of the semi-dense map with the most recent events in TS, as they are both involved in the scene edges. For mapping, a forward-projection method is applied to reproject each pixel in the reference TS to the stereo TS to retrieve their depths by optimizing the stereo temporal consistency, which is enhanced by a depth fusion strategy from previous depth estimation and its neighborhoods. \cite{s23041998} proposes a selection strategy on the support of the semi-dense map to remove redundant depth points and reduce the computation overhead.

\subsection{Multi-sensor Method}
Event-image alignment methods are the multi-sensor methods with additional images, which are discussed in Sec. \ref{subsec:event_image_align}. In this subsection, multi-sensor methods are introduced except event-image alignment methods.

\textbf{RGB-D Sensor.} 
\cite{Gallego2015EventbasedCP, 8094962, 8794255} construct a photometric 3D map from the RGB-D sensor and estimate the camera motion with events. However, this imposes a huge computation burden for pre-processing.
To alleviate this problem, DEVO \cite{9811805} augments event camera with the depth sensor to build an accurate 3D local depth map, which is less influenced by noisy events in the mapping module with respect to the event-based 3D map.

\textbf{IMU Sensor.}
\cite{s23041998} utilizes the IMU pre-integrated algorithm \cite{leutenegger2013keyframe, 7557075} to fuse the IMU data with time surface, which prevents the degeneration of ESVO \cite{9386209} when few events are generated.

\subsection{Discussion}
Direct methods offer an alternative to feature-based methods by not relying on the accuracy of feature detection. Instead, they exploit either spatio-temporal information or the brightness relationship in event data for localization and mapping. Specifically, event-image alignment methods align events to intensity values, while event representation-based alignment methods align with edge patterns depicted in the event representation accumulation. Direct methods are well-suited for event data and demonstrate strong performance in event-based VO. This is due to two key factors. Firstly, events are triggered by moving edges, which act as the pixel selection with strong image gradients on the image plane. Secondly, the high temporal resolution of event data results in a small displacement between two event frames, which is necessary for direct methods to obtain an optimal solution.

% -------------------------------------------------------------------------------------------------------------------------------- %
\section{Motion-compensation Method}
% -------------------------------------------------------------------------------------------------------------------------------- %
\label{sec:motion}

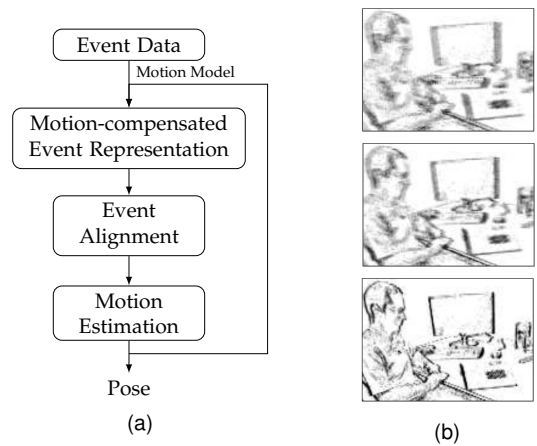
\begin{figure}
    \centering
    \subfloat[]{
        \resizebox{0.4\columnwidth}{!}{
        \begin{tikzpicture}[node distance=1cm]
            \node [process, anchor=north] (event) {Event Data};
            \node [process, below of=event, yshift=-0.5cm, text width=3.5cm] (e-rep) {Motion-compensated Event Representation};
            \draw [flow line] (event) -- coordinate[pos=0.5] (@e) node[anchor=west,yshift=0.2cm,font=\scriptsize]{Motion Model} (e-rep);
    
            \node [process, below of=e-rep, yshift=-0.4cm] (align) {Event Alignment};
            \node [process, below of=align, yshift=-0.4cm] (pose) {Motion Estimation};
            \draw [flow line] (e-rep) -- (align);
            \draw [flow line] (align) -- (pose);

            \node [below of=pose, yshift=-0.2cm] (out) {Pose};
            \draw [flow line] (pose) -- coordinate[pos=0.4] (@p) (out);
            \draw [flow line] (@p) -| ($(pose.east)+(1.0,0.0)$) |- (@e) -- (e-rep);
        \end{tikzpicture}
        }
    }
    \hfil
    \subfloat[]{
        \begin{tabular}[b]{c}
             \includegraphics[width=0.13\textwidth]{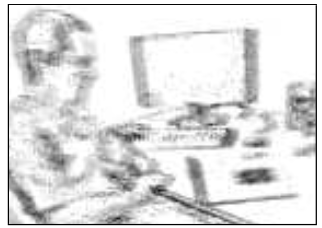} \\
             \includegraphics[width=0.13\textwidth]{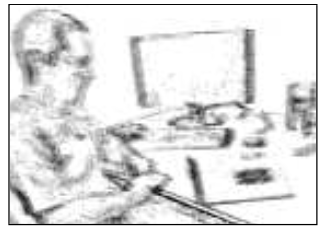}\\
             \includegraphics[width=0.13\textwidth]{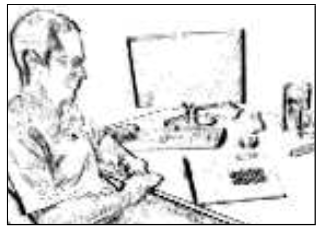}
        \end{tabular}
    }
    \caption{The flow diagram in (a) describes motion-compensation methods in camera tracking, which estimate camera motions by iteratively maximizing event alignment in event accumulations to produce sharp edges from top to bottom in (b). Figure (b) is adapted from \cite{8954483}.}
    \label{fig:motion_compensation_method}
\end{figure}

Motion-compensation methods are built on event alignment, which employs the event frame as the primary event representation, and may be subject to motion blur across an extended temporal window. Motivated by this observation, motion-compensation methods aim to estimate camera motion parameters by optimizing the event alignment in the motion-compensated event frame, leading to the acquisition of sharp images, as shown in Fig. \ref{fig:motion_compensation_method}. However, under certain circumstances, motion-compensation methods may converge towards an undesirable outcome, resulting in event collapse, where events are accumulated into a line or a point within the event frame. In this regard, we classify motion-compensation methods into three categories: 1) the contrast maximization (CMax) method, 2) the dispersion minimization (DMin) method, and 3) the probabilistic alignment method. In this section, we present an overview of representative methods in each category and discuss the event collapse phenomenon.

\subsection{Contrast Maximization}
\label{subsec:cmax}
The CMax framework \cite{8578505} aligns event data triggered from the same scene edges by maximizing the edge strengths in the IWE. The CMax method initially warps a sequence of events into a reference frame using candidate motion parameters, following which the contrast (\textit{i.e.}, variance) of the IWE is maximized to enhance the edge strengths and estimate event camera motion. In \cite{8954483}, various reward functions used to measure the image sharpness and dispersion of the IWE are examined, with experimental results demonstrating that functions such as variance, gradient magnitude, and Laplacian are the most effective. \cite{9454404} proposes a method to reduce drift errors that accumulate with integration in the CMax framework by globally aligning events. Furthermore, globally optimal CMax methods \cite{10.1007/978-3-030-58574-7_4, 9329204, 9156854} are proposed to obtain the optimal solution by deriving the upper and lower bounds of several reward functions and applying the branch-and-bound algorithm for global optimization.

The CMax framework can not only be applied on the camera motion estimation, but also be utilized in the depth estimation task. \cite{8578505} first warps events to the reference frame using candidate depth parameters, and maximizes the contrast by varying the depth parameters. Furthermore, EMVS \cite{10.1007/s11263-017-1050-6} adopts the CMax method in the 3D space to reconstruct event-based 3D map. It first constructs a discrete voxel grid, namely disparity space image (DSI), and back-projects events to 3D rays using camera poses. The ray density in each voxel is then computed as the number of ray intersections and the scene structure is recovered by maximizing the ray densities. Furthermore, the Volumetric Contrast Maximization method \cite{s22155687} computes the volumetric ray density in each voxel as the spatial point-to-line distance between the voxel center and the 3D rays.

\subsection{Dispersion Minimization}
\label{subsec:dmin}
In contrast to the CMax framework, DMin methods \cite{10.1007/978-3-030-58558-7_10, 9625712} utilize the camera motion model to warp events into a feature space rather than reprojecting them into a 2D reference frame. Within the feature space, DMin methods apply entropy loss on the warped events to minimize the average events dispersion, which strengthens edge structures. The entropy loss is computed using the Potential energy and the \textit{Sharma-Mittal} entropy \cite{Gupta1976OnNM}, but this is more computationally expensive than CMax due to the use of a multivariate Gaussian kernel. To address this issue, DMin methods apply a truncated kernel function-based convolution to the feature vector, resulting in a linear computational complexity with the number of events. Additionally, \cite{9625712} develops an incremental version of the DMin method that optimizes the measurement function in its spatio-temporal neighborhood for each incoming event.

\subsection{Probabilistic Model}
\label{subsec:prob_model}
The probabilistic model method \cite{Gu_2021_ICCV} aims to measure the entropy of a set of events at each pixel independently, as opposed to the DMin method using the pairwise entropy measurement. The pixelwise independence assumption leads to a simple theoretical formulation and increases the computational efficiency. To achieve this, it proposed to transfrom an event stream in to an IWE and evaluate the likelihood of event data belonging to the same scene point. The count of warped events at each pixel is modeled as a Poisson random variable and the probability of the entire warped events is formulated as a a collection of independent Poisson point process, \textit{i.e.}, the spatio-temporal Poisson point process (ST-PPP) model \cite{moller2003statistical}. The camera motion parameters are then estimated by maximizing the probability of the ST-PPP model. However, the hyper-parameters in this model are environment-specific, requiring re-measurement upon transitioning between different scenes or event cameras.

\subsection{Event Collapse}
\label{subsec:event_collapse}
The event collapse is a common issue in motion-compensation methods, where events are mapped to a single point or line in the target space. Due to the event collapse, motion-compensation methods may converge to a local optimum with a larger value than the desired solution, as the objective function of the optimization framework is designed for scenarios where event collapse does not occur. To mitigate this issue, a simple and effective strategy is to initialize motion parameters close to the optimal solution \cite{8578505}. \cite{9625712} proposes a data preprocessing procedure (\textit{i.e.}, whitening events) to decorrelate data dimensions and enforce covariance constancy, which helps prevent solution degeneration. Furthermore, two metrics, namely divergence and area-based deformation, are proposed in \cite{s22145190}. These metrics are used to quantify event collapse and design new objective functions with corresponding regularizers.

\subsection{Discussion}
Motion-compensation methods take the advantage of events accumulated over a long temporal window, allowing for the preservation of long-term edge patterns and more robust estimation of camera poses. Furthermore, motion-compensation methods are capable of utilizing timestamps of event data for modeling camera motions. Despite these advantages, current motion-compensation methods suffer from the event collapse in a broad range of camera motions. This issue causes the optimization process to converge to an undesired solution with a larger value than the intended one. As a result, the event collapse problem restricts motion-compensation methods in cases of rotational camera motion.

% -------------------------------------------------------------------------------------------------------------------------------- %
\section{Deep Learning Method}
% -------------------------------------------------------------------------------------------------------------------------------- %
\label{sec:dl}

\begin{figure}
    \centering
        \begin{tikzpicture}[node distance=1cm]
            \node [process, anchor=north] (event) {Event Data};
            \node [process, below of=event, yshift=-0.4cm, text width=2.3cm] (e-rep) {Event Representation};
            \draw [flow line] (event) --  (e-rep);
    
            \node [process, below of=e-rep, xshift=-1.1cm, yshift=-0.4cm, text width=1.35cm] (depth) {Depth Network};
            \node [process, below of=e-rep, xshift=1.1cm, yshift=-0.4cm, text width=1.35cm] (pose) {Pose Network};
            \draw [flow line] (e-rep) -- (depth);
            \draw [flow line] (e-rep) -- (pose);
    
            \node [process, below of=e-rep, yshift=-1.5cm, draw=orange, text=orange] (flow) {Optical Flow};
            \draw [flow line, orange] (depth) -- (flow);
            \draw [flow line, orange] (pose) -- (flow);
            
            \node [below of=depth, xshift=-1.0cm, yshift=-0.13cm, text=blue] (depth-gt) {Depth};
            \node [below of=pose, xshift=1.0cm, yshift=-0.13cm, text=blue] (pose-gt) {Pose};
            \draw [flow line, blue] (depth) -- (depth-gt);
            \draw [flow line, blue] (pose) -- (pose-gt);
        \end{tikzpicture}
    \caption{Diagrams of two typical types of deep learning methods. Self-supervised methods estimate depth and camera pose to recover optical flow as the training signal (in orange), while supervised methods estimate the depth and camera pose with ground truth separately (in blue).}
    \label{fig:deeplearningevent}
\end{figure}
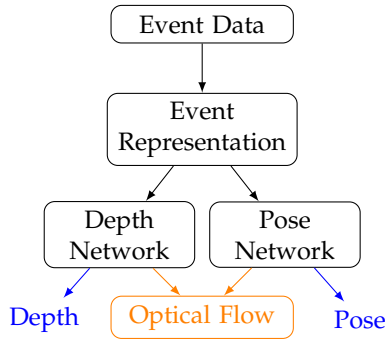

In recent years, deep learning methods have demonstrated considerable potential in vSLAM algorithms \cite{8100183, 7989236, yang20d3vo, 9812347, Zhao2022JPerceiverJP, 10.1007/978-3-031-19839-7_9}. However, the event camera generates asynchronous and sparse event data, making them difficult to process using conventional deep neural networks (DNNs) such as multi-layer perceptrons (MLPs), CNNs, and RNNs. To process event data, current DNNs require conversion to the event-frame based representation \cite{9341224} or voxel grid \cite{8953979}, while SNNs can directly process event data without any pre-processing. In this section, we present a comprehensive review of the research progress in event-based deep learning methods, which are categorized into self-supervised and supervised learning methods (Fig.~\ref{fig:deeplearningevent}).

\subsection{Self-supervised Learning Method}
\label{subsec:dl_unsupervised}
Self-supervised learning methods do not require access to ground truth camera poses and depth values during training. Instead, they rely on supervisory signals derived from temporal dynamics information, such as photometric consistency, by back-warping adjacent frames using depth and pose predictions from the DNNs under the multi-view geometric constraint. Fig. \ref{fig:deeplearningevent} illustrates a typical self-supervised framework \cite{8100183}. In \cite{9341224}, event data is first converted into event frames which are fed into the evenly-cascaded convolutional network that evenly resizes the features to deal with the sparsity, to predict the camera motion and depth. Then, events in consecutive neighboring frames are back-warped to the middle frame to construct the error measurement for camera motion and depth estimation. \cite{8953979} adopts an U-Net-like structure and improves upon \cite{9341224} in two aspects, namely, the event representation and the loss function. First, it proposes the DEV by interpolating event data in a discrete 3D spatio-temporal space, followed by two shared-weights depth networks to predict the disparities from a stereo input pair and a pose network to predict the camera pose. Inspired by the CMax framework, the optical flow is estimated from the camera pose and depth to warp the event into the reference frame and leverage the contrast of the event frame as the training signal. 

\subsection{Supervised Learning Method}
\label{subsec:dl_supervised}
Supervised learning methods aim to minimize the errors between predicted poses and depths with their ground truth. In \cite{9025620}, camera poses are regressed from sequences of event data by utilizing a CNN to extract features from the event frame, followed by the combination of the current information with motion history using a stacked spatial LSTM. Nonetheless, standard CNNs are limited to processing accumulated events, which leading to the same latency as frame-based methods. To mitigates this limitation,  EventNet \cite{sekikawa2019eventnet} treats event data as the point cloud and adopts a variant of PointNet \cite{Qi2016PointNetDL} with an recursive processing module to handle the event data asynchronously.

To leverage the temporal information, E2Depth \cite{9320311} introduces a U-Net architecture with ConvLSTM to predict dense monocular depth from a continuous stream of event data. It first encodes the DEV into a compact representation, which is subsequently fed into the ConvLSTM model to predict a normalized logarithmic depth map. The network is optimized by minimizing both the scale-invariant loss and multi-scale scale-invariant gradient matching loss, which promote smooth depth changes and enforce depth discontinuities. However, the ConvLSTM model is likely to lose the long-term dependency between events. To address this issus, \cite{10205350} proposes a novel hierachical neural memory network to encode the local and dynamic information quickly at low-level memory and memorize the global information from a history of events at high-level memory.

The standard DNN models are restricted to process image-like data at a fixed rate and can only predict a camera pose for a group of event data in each event accumulation. The SNN proposed in \cite{DBLP:conf/icra/GehrigSMS20} can directly process each event data without any preprocessing procedure and perform continuous-time camera pose regression through the spike-generation mechanism. The SNN processes an event, referred to as a spike, at a time and updates the states in the input layer. If the state value in current layer exceeds a pre-defined threshold, a new spike is generated and passed to the next layer. Since the spike-generation mechanism is non-differentiable and an additional temporal dimension is considered, the training process of SNN is extremely difficult. To mitigate this issue, different training paradigms are developed, including conversion from shadow ANNs to SNNs \cite{NIPS2015_10a5ab2d, 7280696, 7727758}, surrogate gradients \cite{Shrestha2018SLAYERSL} and spike-timing-dependent plasticity \cite{Diehl2015UnsupervisedLO}.

\subsection{Multi-sensor Method}
Deep learning methods focus on the incorporation with additional image sensors, while other sensors are still under investigated. In this section, we introduce deep learning methods with both event and frame-based camera.

\textbf{Image Sensor.} First of all, several methods \cite{9359329, 10.1007/978-3-031-20071-7_5} extract features from both event data and images to enhance the performance of vSLAM tasks. The recurrent asynchronous multi-modal (RAM) approach \cite{9359329} proposes a multi-modal model that integrates event data and images by asynchronously updating the internal state of an ConvGRU \cite{8296851}. On the other hand, \cite{10.1007/978-3-031-20071-7_5} builds the features association between event data and images in stereo setting with different levels of features extracted from the FPN. Secondly, self-supervised depth estimation with supervised signals constructed from images has also been investigated in \cite{8968223}, which predicts the ratio between the height above the ground plane and the depth in the camera frame. The depth and height are then extracted from this ratio using a given ground plane transformation. During training, it estimates the optical flow with the predicted height-to-depth ratio and establish photometric losses between two consecutive images. Thirdly, event-based models often learn the edge features while frame-based models can capture the detailed structural information. To leverage the merit of frame-based model, \cite{Wang2021DualTL} proposes a dual transfer learning method: 1) a shared-weights encoder between event end-task learning and event-to-image translation, which is capable to capture the visual structural information; and 2) a feature-level and label-level prediction transfer learning module to further enhance the event end-task learning. Last but not least, Mixed-EF2DNet \cite{10160605} fuse a event-based voxel grid list with the predicted optical flow from frame-based pre-trained RAFT \cite{Teed2020RAFTRA} to fuse a long temporal windows of event data. Then, the voxel grid list and flow features are fed into the depth network to predict contextual depth maps. Finally, the depth map of central voxel grid is estimated by the weighted contextual depth maps to enhance the robustness.

\subsection{Discussion}
Deep learning methods have the capability to capture the non-linear characteristics of event camera, handle noises and outliers, and establish the complex mapping between event data and robot states. However, the effectiveness of these methods relies heavily on having a large amount of training data, and they provide no guarantees of optimality or generality in different environments. SNNs, a special type of DNNs, are well-suited for event-based vSLAM algorithms due to their ability to directly process asynchronous, irregular event data at extremely low power consumption. However, there are several drawbacks to SNNs, such as the spike vanishing problem in deep layers and the notoriously difficult training process, hindering their generalization to a stacked hierarchical model.

% -------------------------------------------------------------------------------------------------------------------------------- %
\section{Performance Comparison}
% -------------------------------------------------------------------------------------------------------------------------------- %
\label{sec:perf}
This section presents the performance comparison of various typical event-based vSLAM algorithms in terms of camera pose and depth estimation. Evaluation of vSLAM algorithms can be conducted qualitatively by subjective visual comparisons and quantitatively using objective metrics. First, we introduce commonly used datasets in SOTA visual odometry systems and some recently proposed datasets, as well as metrics for measuring the performance of camera pose and depth estimation. We then provide a comprehensive evaluation of several vSLAM tasks, including depth estimation, ego-motion estimation, and visual odometry, by collecting and analyzing results from existing studies.

\subsection{Experiment Setting}
\subsubsection{Dataset}
One early dataset \cite{doi:10.1177/0278364917691115} provides several sequences of a hand-held event camera capturing an indoor environment, with ground truth camera poses captured by a motion-capture system. Another dataset, the \textit{rpg} dataset \cite{10.1007/978-3-030-01246-5_15}, uses hand-held stereo event cameras to record an indoor environment. However, both these datasets are limited to low-speed camera motion in small-scale indoor environments. The MVSEC dataset \cite{Zhu2018TheMS} is the first large-scale stereo event camera dataset that leverages a hexacopter to capture more complex scenarios in both indoor and outdoor environments and a driving car to record outdoor day and night scenarios. On the other hand, the \textit{vicon} dataset proposed in \cite{Guan2022PLEVIORM} includes two event cameras with different resolutions that are used to record HDR scenarios with very low illumination, strong illumination changes, or aggressive motion. Recently, several advanced event-based vSLAM datasets \cite{8793887, klenk2021tumvie, 9879881, 9809788, 9664374} have been publicly released, featuring complex settings.

\subsubsection{Metrics}
For evaluating the performance of vSLAM algorithms in terms of camera pose estimation, the absolute trajectory error (ATE) and the relative pose error (RPE) are two commonly used metrics \cite{6385773}. ATE measures camera poses with respect to the world reference, while RPE measures the relative camera poses between two consecutive frames. The translational error in ATE, also known as the positional error, is computed as the Euclidean distance between the estimated and ground truth camera poses. The rotational error in ATE, also known as the orientation error, is computed as the geodesic distance in the rotation group \textit{SO}(3). Similarly, RPE measures the translational and rotational errors between all camera pose pairs with the same time interval. ATE provides a single metric to assess the long-term performance of the VO system, while RPE reflects the local consistency in the relative camera poses. In our evaluation, we employ the positional and orientation errors in ATE to compare the performance of existing works. It is also noteworthy that some studies calculate the positional error with respect to the mean scene depth \cite{8094962} or the total traversed distance \cite{Guan2022PLEVIORM}, which ensures that the error measurement is invariant to the scale of the scene or trajectory.

In addition, \cite{8953979} suggests using three error metrics, namely ARPE, ARRE, and AEE, to evaluate the estimated translational vector and rotational matrix. Specifically, ARPE and AEE quantify the position and orientation differences between two translational vectors, respectively, while ARRE measures the geodesic distance between two rotational matrices.

Despite these above metrics, average linear and angular velocity errors \cite{9454404, 9625712} can also serve as alternative metrics for evaluating camera pose estimation, as the camera poses can be represented as a function of linear and angular velocities with respect to time.

Regarding depth estimation, the average depth error at several cutoffs up to fixed depth values is commonly used to evaluate the performance of event-based depth estimation methods. This metric enables comparisons of different methods across diverse scales of 3D maps.

\subsection{Performance Evaluation of State-of-Art Methods}
We first evaluate the tracking performance of representative event SLAM systems, including multi-sensor methods. For mapping, we report the results of DL methods, since most geometric event SLAM systems only provide qualitative analysis, which is difficult for quantitative comparison.

\subsubsection{Camera Poses Estimation}
We first evaluate the performance of motion-compensation methods on rotational sequences \cite{doi:10.1177/0278364917691115} by measuring the Root Mean Square (RMS) of angular velocity errors from \cite{Gu_2021_ICCV}, as presented in Tab. \ref{tab:angular_motion}. CMax \cite{8578505} exhibits good performance for the 3-DoF rotational motion of event cameras, with the lowest time complexity among the evaluated methods. DMin \cite{10.1007/978-3-030-58558-7_10} extends CMax to high-dimensional feature spaces and applies entropy minimization to the projected events, resulting in an improvement in the performance of approximately 20\%. However, DMin is computationally expensive. To address this issue, approximate DMin uses a truncated kernel to balance performance and efficiency. ST-PPP \cite{Gu_2021_ICCV} offers an alternative solution by employing a probabilistic model, achieving the highest performance among the evaluated methods, with a 39\% improvement in the \textit{shapes} sequence.

\begin{table}[h]
    \centering
    \caption{The angular velocity error (degrees/s) of motion-compensation methods on rotational sequences \cite{doi:10.1177/0278364917691115}.}
    \label{tab:angular_motion}
    \begin{tabular}{ c | c | c | c | c }
        \hline
         & boxes & poster & dynamic & shapes \\
        \hline
        CMax \cite{8578505} & 9.08 & 13.45 & 7.13 & 55.87 \\
        DMin \cite{10.1007/978-3-030-58558-7_10} & 7.06 & 10.86 & 5.39 & 42.22 \\
        Approx. DMin \cite{10.1007/978-3-030-58558-7_10} & 7.81 & 12.36 & 6.19 & 55.44 \\
        ST-PPP \cite{Gu_2021_ICCV} & 6.73 & 10.37 & 5.19 & 25.89 \\
        \hline
    \end{tabular}
\end{table}

Then, we assess the performance of both deep learning and motion-compensation methods on the \textit{outdoor\_day 1} sequence by measuring the APRE, ARRE, and AEE metrics with the results in \cite{9625712}. As shown in Tab. \ref{tab:deep_motion}, DMin \cite{9625712} performs well on the \textit{outdoor\_day 1} sequence by minimizing the dispersion of back-projected events in 3D space. Interestingly, approximate DMin achieves approximately 20\% better performance than standard DMin while also reducing time complexity. However, the online version of DMin performs worst since it operates on an event-by-event basis. Deep learning methods outperform motion-compensation methods, \textit{e.g.}, \cite{9341224} achieves the best performance by estimating 5-DoF ego-motion with a known scaled factor.

\begin{table}[h]
    \centering
    \caption{APRE (degrees), ARRE (radians), AEE (m/s) of deep learning and motion-compensation methods on the MVSEC dataset.}
    \label{tab:deep_motion}
    \begin{tabular}{ c | c  c  c }
        \hline
         & \multicolumn{3}{c}{outdoor day1}  \\
         & APRE & ARRE & AEE \\
        \hline
        DMin \cite{9625712} & 9.41 & 0.01298 & 0.92  \\
        Approx. DMin \cite{9625712} & 8.04 & 0.01524 & 0.83  \\
        Online DMin \cite{9625712} & 13.98 & 0.01129 & 1.29  \\
        \hline
        Zhu \textit{et al.} \cite{8953979} & 7.74 & 0.00867 & - \\
        Ye \textit{et al.} \cite{9341224} & 3.98 & 0.00267 & 0.70  \\
        \hline
    \end{tabular}
\end{table}

\begin{table*}[ht]
    \centering
    \caption{Positional error (Pos., cm) and orientation error (Ori., degrees) of the ego-motion estimation on the boxes sequence \cite{doi:10.1177/0278364917691115} and pipe sequence \cite{8094962}.}
    \label{tab:direct_ego}
    \begin{tabular}{ c | c c | c c | c c | c c | c c }
        \hline
         & \multicolumn{2}{c |}{boxes 1} & \multicolumn{2}{c |}{boxes 2} & \multicolumn{2}{c |}{boxes 3} & \multicolumn{2}{c |}{pipe 1} & \multicolumn{2}{c}{pipe 2} \\
         & Pos. & Ori. & Pos. & Ori. & Pos. & Ori. & Pos. & Ori. & Pos. & Ori. \\
        \hline
        Gallego \textit{et al.} \cite{8094962} & 5.08 & 2.51 & 4.04 & 2.18 & 5.47 & 2.82 & 10.96 & 2.90 & 15.26 & 4.68 \\
        Bryner \textit{et al.} \cite{8794255} & 4.74 & 1.86 & 4.46 & 2.10 & 5.05 & 2.39 & 10.23 & 2.13 & 11.29 & 4.02 \\
        \hline
    \end{tabular}
\end{table*}

\begin{table*}[ht]
    \centering
    \caption{Positional error (Pos., cm) and orientation error (Ori., degrees) of the frame-based and event-based VO methods on the \textit{rpg} dataset \cite{10.1007/978-3-030-01246-5_15}.}
    \label{tab:event_vo}
    \begin{tabular}{ c | c  c | c   c | c   c | c   c }
        \hline
         & \multicolumn{2}{c |}{rpg\_bin} & \multicolumn{2}{c |}{rpg\_boxes} & \multicolumn{2}{c |}{rpg\_desk} & \multicolumn{2}{c}{rpg\_monitor} \\
        & Pos. & Ori. & Pos. & Ori. & Pos. & Ori. & Pos. & Ori. \\
        \hline
        ORB-SLAM2 \cite{7946260} (stereo) & 0.7 & 0.58 & 1.6 & 4.26 & 1.8 & 2.81 & 0.8 & 3.70 \\
        ORB-SLAM2 \cite{7946260} & 2.4 & 0.84 & 3.9 & 2.39 & 3.8 & 2.52 & 3.1 & 1.77 \\
        DSO \cite{7898369} & 1.1 & 2.12 & 2.0 & 2.14 & 10.0 & 63.5 & 0.9 & 0.33 \\
        DSO+ \cite{7898369} & - & - & - & - & 1.6 & 1.80 & 2.1 & 1.54 \\
        \hline
        EVO \cite{7797445} & 13.2$^*$ & 50.26$^*$ & 14.2$^*$ & 170.36$^*$ & 5.2 & 8.25 & 7.8 & 7.77 \\
        USLAM \cite{Vidal2017UltimateSC} & 7.7 & 7.18 & 9.5 & 8.84 & 9.8 & 32.46 & 6.5 & 7.01 \\
        ESVO \cite{9386209} & 2.8 & 7.61 & 5.8 & 9.46 & 3.2 & 7.25 & 3.3 & 2.74 \\
        EDS \cite{9879881} & 1.1 & 0.99 & 2.1 & 1.83 & 1.5 & 1.87 & 1.0 & 0.60 \\
        \hline
        \multicolumn{9}{l}{+ refers to DSO \cite{7898369} method using the reconstructed images from E2VID \cite{8946715}.} \\
        \multicolumn{9}{l}{$^*$ indicates that method failed after completing at most 30$\%$ of the sequence.} \\
        \multicolumn{9}{l}{- indicates failure after initialization, completing less than 10$\%$ of the sequence.} \\
    \end{tabular}
\end{table*}

\begin{table*}[ht]
    \centering
    \caption{Positional error of the frame-based and event-based VIO methods on the \textit{vicon} dataset \cite{Guan2022PLEVIORM}.}
    \label{tab:evio}
    \begin{tabular}{ c | c  c  c  c | c  c | c  c | c  c | c }
        \hline
         & \multicolumn{4}{c |}{vicon hdr} & \multicolumn{2}{c |}{vicon darktolight} & \multicolumn{2}{c |}{vicon lighttodark} & \multicolumn{2}{c |}{vicon dark} & vicon aggressive hdr \\
        & 1 & 2 & 3 & 4 & 1 & 2 & 1 & 2 & 1 & 2 &  \\
        \hline
        VINS-MONO \cite{Qin2017VINSMonoAR} & 0.96 & 1.60 & 2.28 & 1.40 & 0.51 & 0.98 & 0.55 & 0.55 & 0.88 & 0.52 & failed \\
        ORB-SLAM3 \cite{9440682} & 0.32 & 0.75 & 0.60 & 0.70 & 0.75 & 0.76 & 0.41 & 0.58 & failed & 0.60 & failed \\
        PL-VINS \cite{Fu2020PLVINSRM} & 0.67 & 0.90 & 0.69 & 0.66 & 0.84 & 1.50 & 0.64 & 0.93 & 0.53 & failed & 1.94 \\
        \hline
        USLAM \cite{Vidal2017UltimateSC} & 2.44 & 1.11 & 0.83 & 1.49 & 1.00 & 0.79 & 0.84 & 1.49 & 3.45 & 0.63 & 2.30 \\
        EIO \cite{9981970} & 0.59 & 0.74 & 0.72 & 0.37 & 0.81 & 0.42 & 0.29 & 0.79 & 1.02 & 0.49 & 0.66 \\
        PL-EVIO \cite{Guan2022PLEVIORM} & 0.17 & 0.12 & 0.19 & 0.11 & 0.14 & 0.12 & 0.13 & 0.16 & 0.43 & 0.47 & 1.97 \\
        \hline
        \multicolumn{12}{l}{Unit: $\%$/m, 0.5 means the average error would be 0.5m for 100m motion.} \\
    \end{tabular}
\end{table*}

Next, we compare two distinct event-image alignment methods, each employing a given 3D photometric depth map, on the \textit{boxes} \cite{doi:10.1177/0278364917691115} and \textit{pipe} sequences \cite{8094962}. We measure the positional errors with respect to the mean scene depth and orientation errors and report the results from \cite{8794255} in Tab. \ref{tab:direct_ego}. \cite{8094962} achieves remarkable performance by utilizing the filter-based method that leverages the photometric relationship between brightness change and absolute brightness. On the other hand, the approach in \cite{8794255} achieves superior performance by aligning two brightness incremental images using least-squares optimization.

Several event-based VO algorithms are evaluated on the \textit{rpg} dataset \cite{9386209} in terms of positional and orientation errors, reported in \cite{9879881}. As shown in Tab. \ref{tab:event_vo}, EVO \cite{7797445} achieves a solid performance in several sequences but fails in some sequences due to its tendency to lose tracking in the case of quick changes in edge patterns. USLAM \cite{Vidal2017UltimateSC} improves the feature-based VO method by fusing additional inertial data and images with events, achieving better performance than EVO. ESVO \cite{9386209} provides more accurate depth estimation from stereo event cameras and outperforms USLAM in camera pose estimation. Nevertheless, it achieves a slightly less accurate performance than frame-based algorithms, such as ORB-SLAM2 \cite{7946260} with bundle adjustment and DSO \cite{7898369}. The event-image alignment algorithm EDS \cite{9879881}, with PBA in the back-end, achieves a comparable performance to DSO. Additionally, DSO also takes the reconstructed image from E2VID \cite{8946715} as input, achieving better performance on the \textit{rpg\_desk} sequence. However, since E2VID cannot reconstruct high-quality images correctly, DSO+ may fail to align two images which results in losing tracks and failing in sequences with complex textures.

Moreover, we also evaluate several event-based VIO methods on the \textit{vicon} dataset \cite{Guan2022PLEVIORM} by measuring the positional errors with respect to the total trajectory length of the ground truth in Tab. \ref{tab:evio}. Event-based VIO methods produce reliable pose estimation, even when frame-based VO and VIO methods failed. Since the images are unable to capture accurate photometric information in dark scenes and aggressive motion scenarios, frame-based methods are likely to fail in constructing accurate data associations, leading to losing tracks. USLAM \cite{Vidal2017UltimateSC} couples event data with IMU data and intensity images, which achieves a slightly lower performance than that of current SOTA frame-based VIO algorithms. EIO \cite{9981970} improves the performance by applying event-corner feature extraction and tracking algorithm and sliding-windows graph-based optimization. Additionally, PL-EVIO \cite{Guan2022PLEVIORM} extends line-based features in event data and point-based features in intensity images to further improve performance, achieving the best performance among both event-based and frame-based VIO methods.

\begin{table*}[ht]
    \centering
    \caption{Average depth error (meters) of the monocular depth estimation methods at different maximum cutoff depths on the MVSEC dataset \cite{Zhu2018TheMS}.}
    \label{tab:mono_depth}
    \begin{tabular}{ c | c  c  c | c  c  c | c  c  c | c  c  c }
        \hline
         & \multicolumn{3}{c |}{outdoor\_day 1} & \multicolumn{3}{c |}{outdoor\_night 1} & \multicolumn{3}{c |}{outdoor\_night 2} & \multicolumn{3}{c}{outdoor\_night 3} \\
        Cutoff & 10m & 20m & 30m & 10m & 20m & 30m & 10m & 20m & 30m & 10m & 20m & 30m \\
        \hline
        MegaDepth \cite{8578316} & 2.37 & 4.06 & 5.38 & 2.54 & 4.15 & 5.60 & 3.92 & 5.78 & 7.05 & 4.15 & 6.00 & 7.24 \\
        MegaDepth+ \cite{8578316} & 3.37 & 5.65 & 7.29 & 2.40 & 4.20 & 5.80 & 3.39 & 4.99 & 6.22 & 4.56 & 5.63 & 6.51 \\
        \hline
        Zhu \textit{et al.} \cite{8953979} & 2.72 & 3.84 & 4.40 & 3.13 & 4.02 & 4.89 & 2.19 & 3.15 & 3.92 & 2.86 & 4.46 & 5.05 \\
        E2Depth \cite{9320311} & 1.85 & 2.64 & 3.13 & 3.38 & 3.82 & 4.46 & 1.67 & 2.63 & 3.58 & 1.42 & 2.33 & 3.18 \\
        RAM \cite{9359329} & 1.39 & 2.17 & 2.76 & 2.50 & 3.19 & 3.82 & 1.21 & 2.31 & 3.28 & 1.01 & 2.34 & 3.43 \\
        \hline
        \multicolumn{13}{l}{+ refers to MegaDepth \cite{8578316} using the reconstructed images from E2VID \cite{8946715}.} \\
    \end{tabular}
\end{table*}

\subsubsection{Depth Estimation}
We evaluate three event-based monocular depth estimation methods using DNNs, and compare their performance to that of a state-of-the-art frame-based method, MegaDepth \cite{8578316}. \cite{8953979} follows SfMLearner \cite{8100183} scheme to predict depth values in an self-supervised manner, while the other two methods, E2Depth \cite{9320311} and RAM \cite{9359329}, rely on ground truth depth values. However, acquiring accurate ground truth depth values is a challenging task in real-world scenarios. Therefore, these supervised methods are trained on synthetic datasets and fine-tuned on real-world data. Specifically, all methods are trained on the \textit{outdoor\_days 2} sequence of the MVSEC dataset \cite{Zhu2018TheMS}.

We report the performance of the above methods on the MVSEC dataset \cite{Zhu2018TheMS} in terms of average depth errors at different maximum cutoff depths, namely 10m, 20m, and 30m, collected from \cite{9320311, 9359329} in Tab. \ref{tab:mono_depth}. The results demonstrate that event-based methods outperform frame-based methods in scenarios with high-speed motion and low-light conditions. In the \textit{outdoor\_night} sequences that are recorded on a moving car at night, MegaDepth experiences reduced accuracy due to motion blur and low dynamic range. Its performance can be improved by applying it to reconstructed images from event streams. The self-supervised method \cite{8953979} performs better than MegaDepth by achieving average depth errors that are 1-2 meters lower. E2Depth's performance is enhanced by utilizing ground truth labels and additional training on a synthetic dataset. RAM, which fuses asynchronous event data with synchronous intensity images, yields further improvements over event-based methods, implying that static features extracted from intensity images have the potential to enhance the performance of event-based methods.

\subsection{Discussion}
Event-based vSLAM algorithms have the potential to be employed in various scenarios. For example, event cameras can be handheld or mounted on mobile devices like moving vehicles, drones, and ground robots, which may travel at various speeds and encounter challenging lighting conditions in both indoor and outdoor environments. However, the wide range of applications and environments in which event vSLAM is deployed implies that no single methodology can satisfy all requirements.

\noindent \textbf{Feature-based methods}
are able to deal with high-speed camera motion by selecting and processing a small subset of event features, which is computationally efficient. Nonetheless, they are not effective in textureless environments. The integration of external IMU data provides a remedy in this case.
\textbf{Direct methods} 
are known for their robustness in textureless environments but their performance is limited in moderate motions.
\textbf{Motion-compensated methods} 
leverage a long-term appearance by accumulating events over a larger temporal window, which ensures robustness in high-speed motion and large-scale environments. However, such methods are often limited to rotational camera motion due to the event collapse phenomenon. In addition, their accuracy also decreases in textureless environments.
\textbf{Deep learning methods}
can provide accurate estimation given proper training data, as demonstrated in driving scenarios. However, their performance depends on large-scale datasets, which can be expensive to collect. Additionally, current models trained on planar motion datasets may not generalize well in more complex camera motion scenarios.

In summary, event-based vSLAM methods are particularly well-suited for scenarios involving high-speed motion, where the computational complexity and temporal resolution of the algorithms significantly affect system performance. Furthermore, these methods are capable of operating effectively in challenging illumination conditions. However, event-based vSLAM methods struggle in textureless environments or slow-motion scenarios where only a few events are generated or during night sequences with flickering lights. Moreover, event-based systems that rely on photometric information under the constant brightness assumption, such as EGM, may not perform well in illumination-changing environments. Lastly, dynamic objects in the view of moving cameras require further investigation in the context of event-based vSLAM methods.

% -------------------------------------------------------------------------------------------------------------------------------- %
\section{Challenges and Future Directions}
% -------------------------------------------------------------------------------------------------------------------------------- %
\label{sec:ch_future}

\subsection{Challenges}
\label{subsec:challenge}
In this section, we outline key challenges that need to be addressed to develop a practical and robust event-based vSLAM system into two categories: intrinsic and extrinsic challenges. Intrinsic challenges pertain to limitations arising from hardware and working principles of event cameras, while extrinsic challenges arise due to challenging environments encountered by event-based vSLAM algorithms.

\subsubsection{Intrinsic Challenges}

\textbf{Motion-variant Edge Pattern.}
In contrast to frame-based images, the appearance of a stream of event data depends on its motion, making event data association challenging in vSLAM algorithms. Specifically, detecting and tracking motion-invariant features becomes difficult when the edge pattern is lost due to its parallel movement with the camera. While intensity images \cite{7758089} and IMU data \cite{Vidal2017UltimateSC} have been employed to mitigate this issue, developing more robust methods for feature detection, tracking, and event alignment in the event domain remains an under-explored area.

\textbf{Sensor Noise.}
Event data inherently contain noise due to various hardware limitations of event cameras, such as timestamp jitter, pixel manufacturing mismatch, and non-linear circuitry effects. These noise sources can increase the drift error in camera poses and depth estimation, which results in reduced accuracy in long-term camera tracking and global map reconstruction. To mitigate this problem, filter-based methods \cite{8094962} have been proposed that use a resilient sensor model. Nevertheless, a more realistic sensor model that accounts for different types of event noise remains an open research question in event-based vSLAM algorithms.

\textbf{Sparsity of Event Data.}
Unlike frame-based cameras that capture full-intensity images, event cameras respond to moving edges and generate a sequence of sparse event data on the image plane. While frame-based methods utilize redundant brightness information to recover a dense 3D map and recognize objects well, event-based methods may not have sufficient edge structure information to reconstruct the entire environment and each object in it, which is critical for robotic navigation. Recently, learning-based methods \cite{9320311} have been proposed to predict dense depth from events and reveal 3D structures where only a few events are triggered, but these methods struggle to handle objects that were not seen during training. Thus, the reconstruction of the entire environment with sufficient support points remains a challenging problem in the event-based vSLAM community.

\textbf{Theoretical Framework.}
Theoretical analysis plays a critical role in providing guarantees for the generality and validation of vSLAM systems. In traditional frame-based vSLAM algorithms, various theoretical tools have been developed, including the Lagrangian duality framework \cite{Carlone2015LagrangianDI}, information-theoretic framework \cite{Zhang2022InformationTheoreticOL}, and those discussed in the survey \cite{7747236}. However, extending these tools to event-based vSLAM systems and developing new theoretical tools that are specific to event sensors, present critical and challenging research endeavors.

\subsubsection{Extrinsic Challenges}

\textbf{Adverse Conditions.}
Despite the benefits of event cameras in HDR environments, certain scenes can degrade the quality of event data, such as night scenes and extreme weather conditions. In particular, the presence of flickering light in a nighttime scenario can result in reduced accuracy of camera pose and depth estimation, as reported in \cite{8953979, 9386209}. This flickering light is often regarded as noise and removed from the raw event data, as in \cite{9812003}. However, modeling the event noise that arises in adverse conditions remains a challenging task for improving the robustness of event-based vSLAM algorithms.

\textbf{Textureless Environments.}
Textureless environments pose a challenge for event-based vSLAM algorithms as high-contrast regions are scarce and trigger few events in such areas. Consequently, event triggers at the boundary lack detailed information about the surrounding environment and provide insufficient support for camera tracking and mapping tasks. Therefore, the development of a robust event-based vSLAM algorithm that can operate effectively in textureless environments remains a challenging problem.

\textbf{Dynamic Environments.} 
In dynamic environments, event-based vSLAM algorithms must detect and track moving or deformable objects and model permanent or temporary changes to update the map accordingly. Several event-based motion segmentation methods have been proposed to detect moving objects in different motions from event data in recent years \cite{8593805, Stoffregen2019EvMS, 9157202}, but none of these methods have yet been adopted in event-based vSLAM systems to simultaneously track objects and the camera.

\subsection{Future Directions}
\label{subsec:future}

\subsubsection{Representation}
\textbf{Event Representation.}
Event-based vSLAM algorithms often transform the event stream into an alternative image-like representation and apply conventional frame-based techniques to process event data synchronously. However, this approach overlooks the sparsity of event data, leading to redundant computation, and assumes a constant position, exacerbating the drift error. In addition, the resulting event frames may lose edge structure over a short temporal window and exhibit motion blur over a long temporal window. Furthermore, when the event camera is stationary, the time surface degenerates. Therefore, developing more robust event representations that fully leverage the asynchronous and sparse nature of event data holds great potential to significantly enhance the performance of current event-based vSLAM systems.

\textbf{Event-based Semantic Map Representation.}
Brightness images provide detailed visual information about an environment, facilitating semantic segmentation and 3D reconstruction via deep learning methods. By contrast, sparse event data may not contain enough information to perform these tasks, and large-scale event-based datasets for such tasks are currently unavailable. To address this challenge, several deep learning approaches have been proposed \cite{9577427, 9025483, 10.1007/978-3-031-19830-4_20} that leverage the favorable characteristics of event cameras in challenging conditions and perform semantic segmentation on event data using prior knowledge from large-scale image datasets. Nevertheless, it is worth further research to employ deep learning methods to generate a complete 3D semantic map from event data.

\textbf{Neural Representation.} NeRF \cite{Mildenhall2020NeRFRS} is a representative neural implicit 3D representation which models a scene through MLP as a continuous function of the scene radiance and volume density learnt from a set of 2D RGB images. Inevitably, NeRF suffers from intrinsic disadvantages of frame-based cameras. To mitigate these issues, recent works \cite{Klenk2022ENeRFNR, Rudnev2022EventNeRFNR, Hwang2022EvNeRFEB} focus on developing event-based NeRFs, reconstructed from event data. These works have experimentally proven that event-based NeRF methods are able to perform better than the frame-based NeRF methods. Future research could further improve the techniques and incorporate it with the camera tracking in event-based vSLAM. Furthermore, 3D Gassuain Splatting \cite{Kerbl20233DGS} achieves real-time rendering, which is a promising future direction for the real-time mapping module in the event-based vSLAM methods.

\subsubsection{Robustness}

\textbf{Event-based Global Optimization.}
In vSLAM algorithms, the accumulation of drift errors in camera poses and 3D points can result in larger errors during long-term camera motion. Event-based vSLAM algorithms also suffer from this issue due to inaccurate data association and sensor noises. While global optimization provides a promising solution to address the above issue, it has been rarely studied in existing event-based vSLAM algorithms, as they primarily focus on local optimization. Recently, PL-EVIO \cite{Guan2022PLEVIORM} presents a loop closure module that matches BRIEF descriptors of event features to refine local camera poses. Similar techniques have proven crucial in frame-based vSLAM algorithms, making it essential to conduct further research in the event domain.

\textbf{Place Recognition and Relocalization.}
In frame-based vSLAM algorithms, it is generally easier and more robust to perform place recognition and relocalization as similar images can be generated for the same location. However, due to the motion-dependent nature of event data, the event camera may generate different event streams for the same location if it moves in different directions. One possible solution to this challenge is to develop motion-invariant feature extraction methods that describe the view from the event camera for place recognition. Subsequently, we can relocalize the camera in a global 3D map using either 2D view recognition or 3D map matching.

\subsubsection{Multi-sensor Setup and Benchmark}

\textbf{Mutli-Sensor for Event-based SLAM.}
Incorporating information from additional sources can enhance the robustness of event-based vSLAM algorithms due to the limited amount of information conveyed by a single event. \cite{9879881} leverages the EGM, which encodes the relationship between absolute brightness and brightness changes, to link the intensity image with the event data. The RAM method \cite{9359329} takes a step further and proposes a novel network structure to integrate asynchronous data from multiple sensors for depth estimation. These multi-sensor methods, which exploit the complementary characteristics of different sensors, present a promising future direction towards more accurate and robust vSLAM systems.

\textbf{Benchmark Datasets.} 
Event-based vSLAM algorithms often use datasets with a wide range of event camera configurations and scenario settings, which makes it challenging to compare their performance. Performance comparison becomes further complicated due to the factors such as sensor noise, contrast threshold, motion pattern, and lighting conditions. Therefore, it is necessary to establish a benchmark dataset that covers diverse challenging scenarios and different sensor suites, to enable a fair and comprehensive performance evaluation of event-based vSLAM algorithms using the same criterion.

\subsubsection{Foundation Model}
\textbf{Foundation Model.} Recently, large-scale models \cite{Ouyang2022TrainingLM, kirillov2023segany} pretrained on massive datasets have achieved remarkable success in both language and vision domains. Event-based camera response to the moving edges which lacks of full information of the whole environment, while the foundation model contains abundant prior knowledge which can better perceive and reconstruct the environment with properties, such as semantics, size and usage. Integrating foundation models into event-based vSLAM systems holds promise for significant performance improvements. Furthermore, active event-based vSLAM, where human guidance can be derived from modern foundation models, offers the potential for faster and more accurate localization and mapping.

% -------------------------------------------------------------------------------------------------------------------------------- %
\section{Conclusion}
% -------------------------------------------------------------------------------------------------------------------------------- %
\label{sec:con}
This paper presents a comprehensive review of the progress made in event-based vSLAM, including the backgound of event cameras and vSLAM algorithm, and the four main types of event-based vSLAM methods, along with performance comparison and the challenges and opportunities. Event cameras have enhanced the robustness of vSLAM algorithms in high-speed scenarios and HDR environments. To unlock the potential advantages of event cameras, some event-based vSLAM methods adopt conventional frame-based methods on an alternative event representation, while others deploy filter-based methods to directly process event data. However, further research is needed to address the sparse, noisy, and motion-variant nature of event data to provide a more robust event-based vSLAM system. Deep learning methods have demonstrated a powerful expressive capacity in conventional vision tasks, and as such, hold the potential for modeling the non-linear characteristics of event data, and advancing the frontier of event-based vSLAM.

% if have a single appendix:
%\appendix[Proof of the Zonklar Equations]
% or
%\appendix  % for no appendix heading
% do not use \section anymore after \appendix, only \section*
% is possibly needed

% use appendices with more than one appendix
% then use \section to start each appendix
% you must declare a \section before using any
% \subsection or using \label (\appendices by itself
% starts a section numbered zero.)
%

% \appendices
% \section{Proof of the First Zonklar Equation}
% Appendix one text goes here.

% you can choose not to have a title for an appendix
% if you want by leaving the argument blank
% \section{}
% Appendix two text goes here.

% use section* for acknowledgment
% \ifCLASSOPTIONcompsoc
%   % The Computer Society usually uses the plural form
%   \section*{Acknowledgments}
% \else
%   % regular IEEE prefers the singular form
%   \section*{Acknowledgment}
% \fi

% The authors would like to thank...

% Can use something like this to put references on a page
% by themselves when using endfloat and the captionsoff option.
\ifCLASSOPTIONcaptionsoff
  \newpage
\fi

% trigger a \newpage just before the given reference
% number - used to balance the columns on the last page
% adjust value as needed - may need to be readjusted if
% the document is modified later
%\IEEEtriggeratref{8}
% The "triggered" command can be changed if desired:
%\IEEEtriggercmd{\enlargethispage{-5in}}

% references section

% can use a bibliography generated by BibTeX as a .bbl file
% BibTeX documentation can be easily obtained at:
% http://mirror.ctan.org/biblio/bibtex/contrib/doc/
% The IEEEtran BibTeX style support page is at:
% http://www.michaelshell.org/tex/ieeetran/bibtex/
\bibliographystyle{IEEEtran}
% argument is your BibTeX string definitions and bibliography database(s)
%\bibliography{IEEEabrv,../bib/paper}
\bibliography{IEEEabrv, ref}
%
% <OR> manually copy in the resultant .bbl file
% set second argument of \begin to the number of references
% (used to reserve space for the reference number labels box)
% \begin{thebibliography}{1}

% \end{thebibliography}

% biography section
% 
% If you have an EPS/PDF photo (graphicx package needed) extra braces are
% needed around the contents of the optional argument to biography to prevent
% the LaTeX parser from getting confused when it sees the complicated
% \includegraphics command within an optional argument. (You could create
% your own custom macro containing the \includegraphics command to make things
% simpler here.)
%\begin{IEEEbiography}[{\includegraphics[width=1in,height=1.25in,clip,keepaspectratio]{mshell}}]{Michael Shell}
% or if you just want to reserve a space for a photo:

% \begin{IEEEbiography}{Kunping Huang}
% Biography text here.
% \end{IEEEbiography}

% if you will not have a photo at all:
\begin{IEEEbiography}
[{\includegraphics[width=1in,height=1.25in,clip,keepaspectratio]{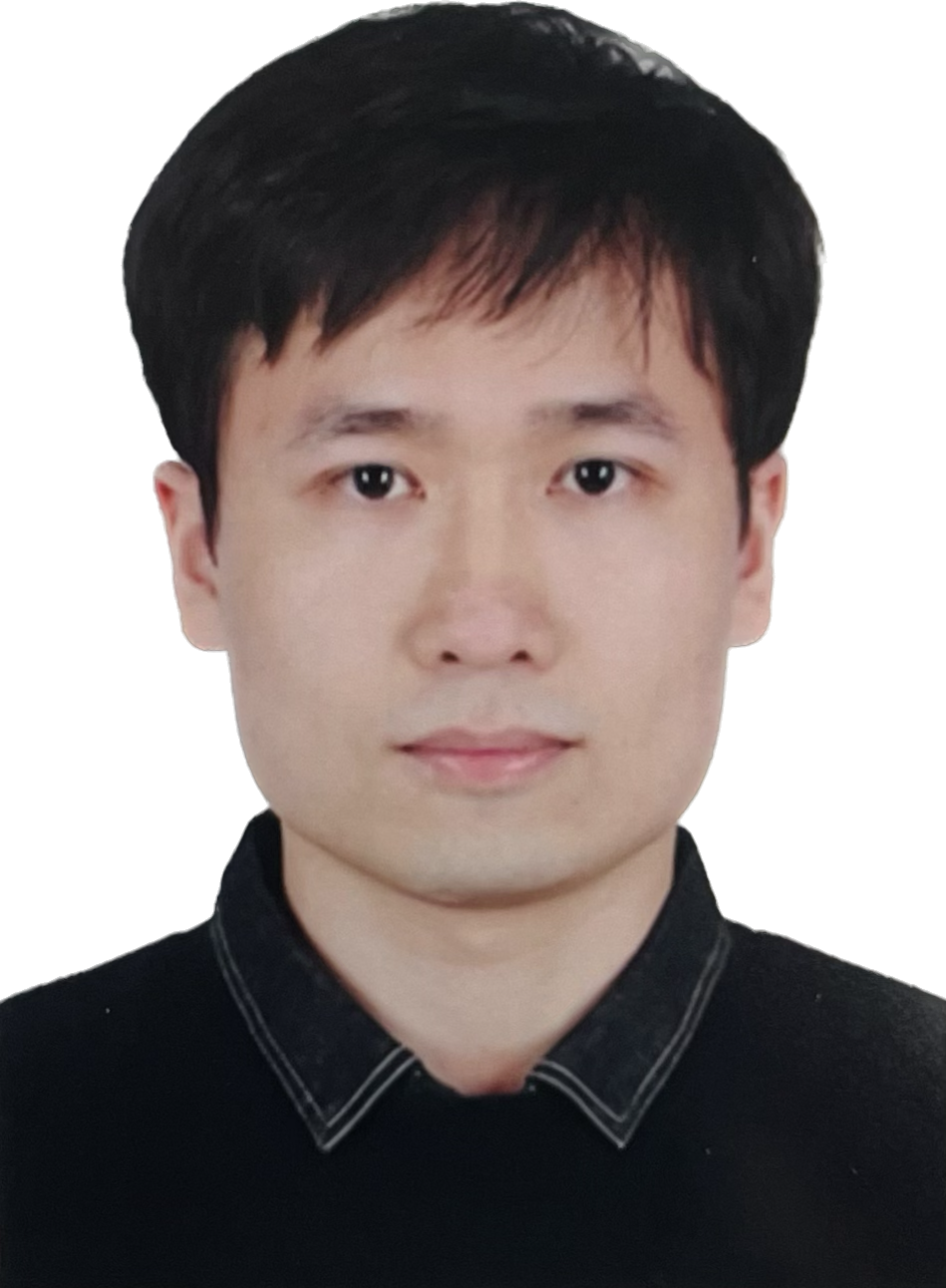}}]{Kunping Huang}
is a PhD student at University of Sydney, supervised by Dacheng Tao and Jing Zhang. He obtained his master's degree at the University of Texas A\&M University in 2020. His research is primarily focused on the development and optimization of vSLAM techniques, with a particular emphasis on utilizing event cameras.
\end{IEEEbiography}

\begin{IEEEbiography}
[{\includegraphics[width=1in,height=1.25in,clip,keepaspectratio]{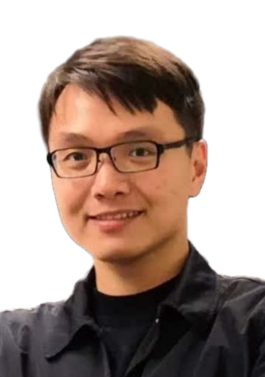}}]{Sen Zhang}  received the B.E. degree in biomedical engineering from Tsinghua University and Master of Philosophy from Hong Kong University of Science and Technology. He is currently a PhD student in the School of Computer Science from the University of Sydney. His research interests include computer vision, depth estimation, and vSLAM. He has published several papers in top-tier conferences and journals including ECCV, IJCV, ICRA, and ACM Multimedia.
\end{IEEEbiography}

\begin{IEEEbiography}
[{\includegraphics[width=1in,height=1.25in,clip,keepaspectratio]{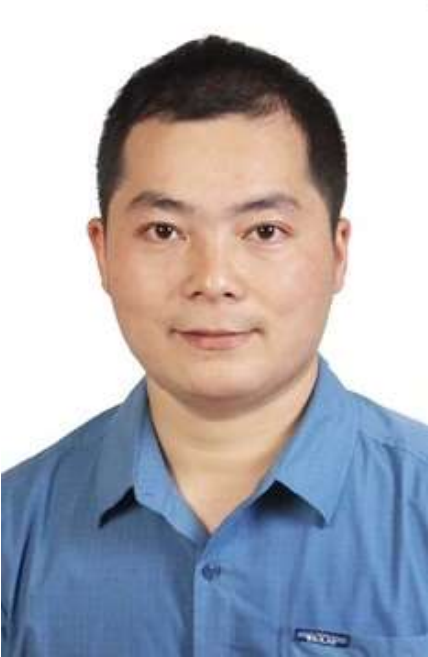}}]{Jing Zhang} (Senior Member, IEEE) is currently a Research Fellow at the School of Computer Science, The University of Sydney. His research interests include computer vision and deep learning. He has published more than 60 papers in prestigious conferences and journals, such as CVPR, ICCV, ECCV, NeurIPS, ICLR, International Journal of Computer Vision (IJCV), and IEEE Transactions on Pattern Analysis and Machine Intelligence (TPAMI). He is a Senior Program Committee Member of AAAI and IJCAI.
\end{IEEEbiography}

\begin{IEEEbiography}
[{\includegraphics[width=1in,height=1.25in,clip,keepaspectratio]{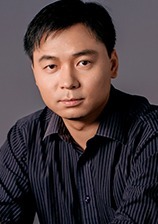}}]{Dacheng Tao} (Fellow, IEEE) is currently a professor of computer science and an ARC Laureate Fellow in the School of Computer Science and the Faculty of Engineering at The University of Sydney. He mainly applies statistics and mathematics to artificial intelligence and data science. His research is detailed in one monograph and over 200 publications in prestigious journals and proceedings at leading conferences. He received the 2015 Australian Scopus-Eureka Prize, the 2018 IEEE ICDM Research Contributions Award, and the 2021 IEEE Computer Society McCluskey Technical Achievement Award. He is a fellow of the Australian Academy of Science, AAAS, ACM, and IEEE.
\end{IEEEbiography}

% insert where needed to balance the two columns on the last page with
% biographies
%\newpage

% \begin{IEEEbiographynophoto}{Jane Doe}
% Biography text here.
% \end{IEEEbiographynophoto}

% You can push biographies down or up by placing
% a \vfill before or after them. The appropriate
% use of \vfill depends on what kind of text is
% on the last page and whether or not the columns
% are being equalized.

%\vfill

% Can be used to pull up biographies so that the bottom of the last one
% is flush with the other column.
%\enlargethispage{-5in}

% that's all folks
\end{document}